%% file: main.tex
\documentclass{article}

\usepackage[preprint]{neurips_2024}

\usepackage{graphicx}%
\usepackage{multirow}%
\usepackage{amsmath,amssymb,amsfonts}%
\usepackage{amsthm}%
\usepackage{mathrsfs}%
\usepackage[title]{appendix}%
\usepackage{xcolor}%
\usepackage{hyperref}
\definecolor{dark-blue}{rgb}{0.15,0.15,0.4}
\definecolor{medium-blue}{rgb}{0,0,0.5}
\hypersetup{
  colorlinks, linkcolor={medium-blue},
  citecolor={medium-blue}, urlcolor={medium-blue}
}
\usepackage{textcomp}%
\usepackage{manyfoot}%
\usepackage{booktabs,arydshln}%
\usepackage{algorithm}%
\usepackage{algorithmicx}%
\usepackage{algpseudocode}%
\usepackage{listings}%
\usepackage{siunitx}
\usepackage{tikz}
\usepackage{tabularx}
\usepackage{float}
\usetikzlibrary{calc, decorations, calligraphy}
\usepackage{subcaption}
\usepackage{wrapfig}
\usepackage[utf8]{inputenc} 
\usepackage[T1]{fontenc}    
\usepackage{bookmark}
\usepackage{cleveref}
\usepackage{url}            
\usepackage{nicefrac}       
\usepackage{microtype}      
\usepackage{physics}
\usepackage{array}

\newcommand{\N}[1]{\mathcal{N}\left(#1\right)} 
\newcommand{\pkernel}[2]{p_{0t} \left( #1 | #2 \right)}
\newcommand{\eye}{\mathbf{I}}

\newcommand{\scoreF}{\mathbf{s}_\theta}

\newcommand{\x}{\mathbf{x}}
\newcommand{\y}{\mathbf{y}}

\newcommand{\w}{\mathbf{w}_{t}}
\newcommand{\dt}{\dd t}
\newcommand{\dw}{\dd \mathbf{w}_{t}}
\newcommand{\f}{f}
\newcommand{\g}{g}

\title{DiffScale: Continuous Downscaling and Bias Correction of Subseasonal Wind Speed Forecasts using Diffusion Models}

\author{%
  Maximilian Springenberg\thanks{These authors contributed equally to this work.} \\
  Applied Machine Learning Group\\
  Fraunhofer Heinrich-Hertz Institute\\
  Berlin, 10587 Germany\\
  \And
  Noelia Otero$^\text{*}$ \\
  Applied Machine Learning Group\\
  Fraunhofer Heinrich-Hertz Institute\\
  Berlin, 10587 Germany\\
  \And
  Yuxin Xue\\
  Applied Machine Learning Group\\
  Fraunhofer Heinrich-Hertz Institute\\
  Berlin, 10587 Germany\\
  \And
  Jackie Ma\\
  Applied Machine Learning Group\\
  Fraunhofer Heinrich-Hertz Institute\\
  Berlin, 10587 Germany\\
}

\makeatletter
\def\adl@drawiv#1#2#3{%
        \hskip.5\tabcolsep
        \xleaders#3{#2.5\@tempdimb #1{1}#2.5\@tempdimb}%
                #2\z@ plus1fil minus1fil\relax
        \hskip.5\tabcolsep}
\newcommand{\cdashlinelr}[1]{%
  \noalign{\vskip\aboverulesep
           \global\let\@dashdrawstore\adl@draw
           \global\let\adl@draw\adl@drawiv}
  \cdashline{#1}
  \noalign{\global\let\adl@draw\@dashdrawstore
           \vskip\belowrulesep}}
\makeatother

\begin{document}

\maketitle

\begin{abstract}
Renewable resources are strongly dependent on local and large-scale weather situations. Skillful subseasonal to seasonal (S2S) forecasts -beyond two weeks and up to two months- can offer significant socioeconomic advantages to the energy sector. This study aims to enhance wind speed predictions using a diffusion model with classifier-free guidance to downscale S2S forecasts of surface wind speed. We propose \textit{DiffScale}, a diffusion model that super-resolves spatial information for continuous downscaling factors and lead times. Leveraging weather priors as guidance for the generative process of diffusion models, we adopt the perspective of conditional probabilities on sampling super-resolved S2S forecasts. We aim to directly estimate the density associated with the target S2S forecasts at different spatial resolutions and lead times without auto-regression or sequence prediction, resulting in an efficient and flexible model. Synthetic experiments were designed to super-resolve wind speed S2S forecasts from the European Center for Medium-Range Weather Forecast (ECMWF) from a coarse resolution to a finer resolution of ERA5 reanalysis data, which serves as a high-resolution target. The innovative aspect of \textit{DiffScale} lies in its flexibility to downscale arbitrary scaling factors, enabling it to generalize across various grid resolutions and lead times -without retraining the model- while correcting model errors, making it a versatile tool for improving S2S wind speed forecasts. We achieve a significant improvement in prediction quality, outperforming baselines up to week 3.
\end{abstract}

\twocolumn

\section{Introduction}\label{sec1}

Subseasonal to seasonal (S2S) forecasting bridges the gap between medium-range weather forecast (up to approximately $10$ days) and seasonal predictions (3–6 months) \citep{Vitart2017}. S2S climate predictions are influenced by the initial conditions of the atmosphere and slowly varying boundary conditions, such as sea surface temperatures, soil moisture, and sea-ice components, which can retain memory from internal processes \citep{White2017, Doblas-Reyes2013}. S2S predictions provide valuable information for a wide range of decision-makers who can benefit from understanding the climate-related risks to optimize resource management and plan ahead \citep{Vitart2017, Vitart2018}. Forecasting subseasonal time scales can be particularly beneficial for the energy sector--for both users and providers--as weather-related risk is a key driver for energy pricing, production, and usage \citep{White2017}. Despite the tremendous effort in improving S2S forecasts, the low forecast skill at S2S time horizons significantly affects the practical utility of subseasonal predictions for policy planners and stakeholders \citep{White2017}. 

To address these systematic errors in physics-based models at the subseasonal scale, recent efforts have focused on demonstrating the potential of machine learning and deep learning methods to enhance S2S forecasting accuracy \citep{Vannitsem2021, Scheuerer2020}. Recently, \cite{Mouatadid2020} proposed a deep learning adaptive bias correction approach to successfully correct the bias of S2S temperature and precipitation in the contiguous U.S. Similarly, \cite{Horat2024} used a CNN-based post-processing method and showed a clear improvement over physical and climatological reference forecasts for global temperature and precipitation with lead times of 14 and 28 days. 

Accurate S2S predictions of key climate variables, such as wind speed, offer valuable insights to the energy industry regarding anticipated renewable energy production and consumption. With the global energy transition towards low-carbon power systems and energy systems becoming increasingly reliant on weather-dependent energy sources, there is an increasing demand for S2S predictions,  beyond short-term weather forecasts \citep{Mariotti2020}. For instance, improving the spatial resolution of wind speed has been shown to significantly enhance the estimation of wind power density \citep{schmidt2024}. Furthermore, the forecasting skill at longer lead times is strongly influenced by the spatial scale of weather events, making higher resolution crucial for accurate S2S forecasts. Consequently, downscaling S2S forecast systems for wind speed can lead to more precise predictions, providing a clearer understanding of wind power generation potential over extended time horizons \citep{Bloomfield2021}. 

Enhancing the spatial resolution of data derived from low-resolution inputs is a key focus across multiple scientific disciplines, especially in climate science, where this process is known as climate downscaling.
Downscaling refers to methods used to refine low-resolution climate forecasts or simulations to higher spatial resolutions, improving their applicability for regional and local-scale analyses. Two main techniques are typically employed: (1) empirical downscaling, which includes statistical and machine learning (ML) methods, and (2) dynamical downscaling, which relies on high-resolution physical models.
The rapid advancement of modern machine learning (ML), including deep learning techniques, has been widely adopted across various climate and weather-related tasks, particularly in post-processing applications like downscaling \citep{koldunov2024, harder2024}. Notable architectures and training strategies, such as convolutional neural networks (CNNs) and generative adversarial networks (GANs) have been successfully applied in climate downscaling \citep{Hohlein2020, Miralles2022, LeToumelin2023, Gao2023}. Diffusion models, in particular, have emerged as a leading approach, demonstrating a strong capability to capture complex spatial and temporal patterns in climate data \citep{mardani2024}. Diffusion models have shown promising results in estimating wind power potential by more effectively preserving the distributional and physical properties of wind speeds \citep{schmidt2024}. This has led to significantly improved downscaling accuracies compared to traditional statistical methods. 

Thus, motivated by the significant advances in image Super-Resolution (SR) shown by diffusion models \citep{song2020score}, we apply a conditional diffusion model to downscale S2S forecasts of surface wind speed. Diffusion models can be applied to SR, due to their support for conditional sampling, which guides the generative process of the downscaled image with a lower resolution version of that image. Typically, downscaling using deep learning methods is handled by fixing a downscaling factor. In this work, we propose \textit{DiffScale}, a diffusion model that generates new spatial information for variable downscaling factors and target resolutions. By embedding the scaling factor as a conditioning input, \textit{DiffScale} achieves super-resolution across multiple scaling factors without requiring retraining.
This novel approach leverages the inherent flexibility of diffusion models, beyond traditional fixed-factor methods to establish a versatile and foundational framework for climate downscaling. 
Conventional models require retraining for each resolution or scaling factor, which is computationally expensive and limits flexibility. In contrast, our approach allows for dynamic downscaling across multiple scales without retraining, enabling greater adaptability in S2S forecasting.
This flexibility not only reduces computational overhead but also addresses practical challenges, such as adapting to varying spatial resolutions and application-specific requirements. By enabling dynamic, multi-factor downscaling within a unified framework, our approach unlocks new possibilities for improving the precision and applicability of S2S forecasts, which have yet to fully harness the potential of modern generative modeling techniques. Our contributions are:
\begin{itemize}
    \item We introduce a principled approach to improving S2S forecasts of surface wind speed across continuous lead times and scaling factors, without relying on auto-regression. Our method leverages the probabilistic nature of sampling from diffusion models to enhance flexibility and accuracy.
    \item We propose \textit{DiffScale}, a diffusion model that generates spatial information for variable downscaling factors and lead times.
    \item Our approach solves multiple downscaling scenarios with a single trained model, reducing computational overhead and addressing challenges in adapting to varying spatial resolutions and application-specific needs.
\end{itemize}

\section{Results}\label{sec2}

In the following, we outline the improved results we were able to obtain from \textit{DiffScale}. We will focus on a qualitative analysis of spatiotemporal results in \Cref{sec:eval-quant} and a quantitative analysis of metrics with respect to lead time in \Cref{sec:eval-qual}. Different configurations of \textit{DiffScale} are listed in \Cref{tab:experiments}. Unless stated otherwise all reported results refer to the lr-ws \& sf configuration.

\begin{table}[b]
    \centering
    \begin{tabularx}{\linewidth}{lXXX}
    \toprule
        Configuration  & Low-Resolution ECMWF Windspeed & Static Fields \hspace{5mm}\vfill (Regional Priors) & Dynamic Fields \hspace{5mm}\vfill  (Weather Variables) \\
    \midrule
        lr-ws \& sf     & \checkmark & \checkmark & -- \\
        sf \& df        & -- &  \checkmark &  \checkmark \\
        lr-ws, sf \& df & \checkmark & \checkmark & \checkmark \\
    \bottomrule
    \end{tabularx}
    \caption{Description of the experimental setup, detailing the input variables and the inclusion of low-resolution ECMWF predictions, static fields, and additional weather variables for each configuration.}
    \label{tab:experiments}
\end{table}

\paragraph{Experiment setup}
\label{sec:expirements}

The proposed \textit{DiffScale} diffusion model can generate samples for continuous lead times and scaling factors. However, to evaluate the method and observe trends, as well as allow for common comparisons, we discretize lead times and scaling factors. Given the size of numerical inputs $\x \in \mathbb{R}^{L\times L}$, we restricted the experiments and evaluations to a set of scaling factors $\alpha \in \left\{ \frac{S}{L\cdot i}
\mid i \in \mathbb{N}^{[1,4]} \right\}$, i.e. scaling factors that result in the fractions $S/4, S/3, S/2$, and $S$ of the maximum resolution $S$. The maximum resolution is determined by the size of the highest resolution counterparts $\y \in \mathbb{R}^{S\times S}, S/4 \geq L$. To simplify readability, we will refer to the resulting fractions of the maximum resolution, rather than the scaling factor.

\begin{figure*}[t]
    \centering
    \includegraphics[width=0.95\textwidth]{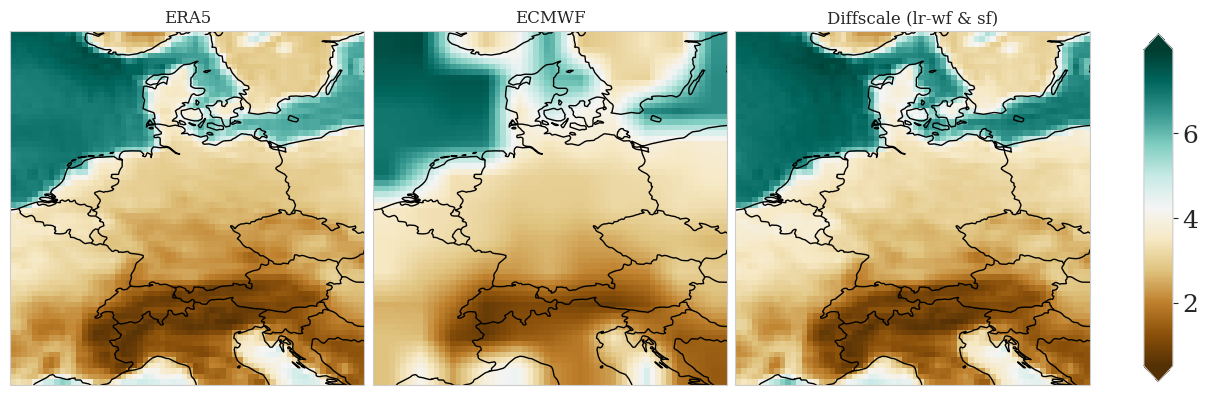}
    \caption{Visual comparison of the ERA5 target, ECMWF S2S and \textit{DiffScale} (lr-ws \& sf) for the finest resolution considered in our method. Displayed is the mean of ws10m obtained over all forecast times.}
    \label{fig:fig1}
\end{figure*}

For our ablation studies, we trained the models with the following conditioning configurations of input data: 1) low-resolution forecast along with static fields (lr-ws \& sf), 2) additional low-resolution weather variables (see \Cref{tab:input_variables} along with static fields, excluding the forecast (sf \& lr-df), and 3) low-resolution forecast along with additional low-resolution weather variables and static fields (lr-ws, sf \& lr-df), as detailed in \Cref{tab:experiments}. \Cref{fig:fig1}  provides a preliminary visual inspection of the results obtained for all configurations (\Cref{tab:experiments}). The models were then evaluated for all configurations, including resolutions and lead times.

\subsection{Quantitative model evaluation}  \label{sec:eval-quant}

\begin{figure*}[t]
    \centering
    \includegraphics[width=0.95\textwidth]{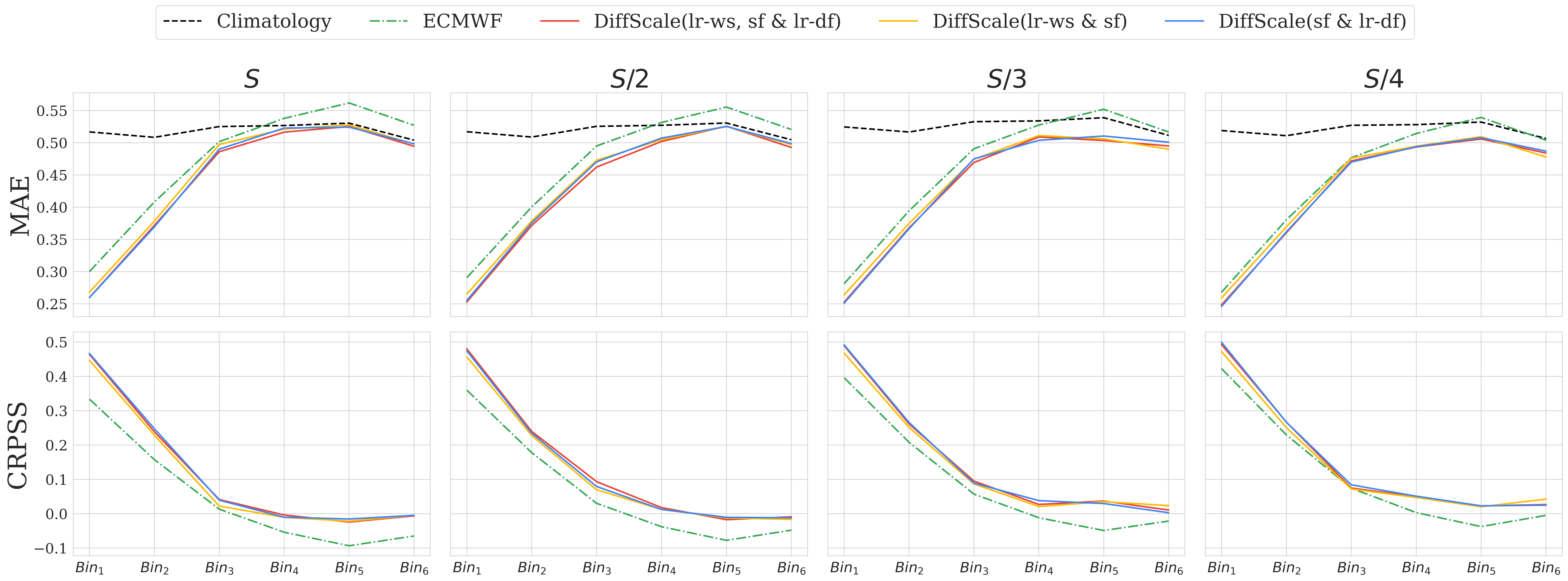}
    \caption{MAE and CRPSS metrics calculated across the different bins for each resolution. The different setups of \textit{DiffScale}'s experiments are shown as colored lines, while the ECMWF S2S model is represented as a dashed and dotted green line. The dashed black line corresponds to climatology.}
    \label{fig:fig_eval}
\end{figure*}

\paragraph{Improved performance} \Cref{fig:fig_eval} depicts the performance of the models across all configuration setups (see \Cref{sec:expirements}), assessed using deterministic MAE and probabilistic CRPSS. The values of those metrics are shown in \Cref{tab:results}. The results demonstrate that \textit{DiffScale} consistently outperforms the baseline ECMWF S2S model across all lead times, suggesting that \textit{DiffScale} is able to capture the dynamics of the ws10, resulting in lower error rates.  The results do not reveal significant differences between the \textit{DiffScale} configurations (see \Cref{tab:experiments}). Overall, the experiment of \textit{DiffScale} (sf \& lr-df), using only low-resolution atmospheric inputs (excluding ws10), shows slightly worse metrics. Interestingly, no significant performance improvement was observed for the experiment with additional inputs (e.g., all inputs included, \textit{DiffScale} (lr-ws, sf \& lr-df) as referred in \Cref{tab:experiments}), compared to the experiment with static fields and low-resolution ws10 as input (\textit{DiffScale} (lr-ws \& sf)), which in general shows higher skill scores across most lead times. We still noticed that the inclusion of additional atmospheric inputs enhances the forecast skill at the longest lead times (i.e. $Bin_5$ and $Bin_6$). However, further analysis is needed to fully understand the influence of atmospheric inputs on forecast performance.

\paragraph{Performance across lead times} 
The MAE exhibits an expected increase with lead time, which becomes particularly noticeable from lead time 3 onward, highlighting the challenge of maintaining accuracy at longer lead times. While \textit{DiffScale} performs strongly at shorter lead times, retaining a good skill until the beginning of week 3 (i.e., $Bin_4$) for all resolutions, it appears to struggle at the highest resolution (Factor S), when compared to the climatological benchmark at week 3. This is evident in the CRPSS decreasing with lead time, with skill relative to climatology diminishing more sharply at finer resolutions (S and S/2) and approaching or dropping below zero at lead time bins 5 and 6. A more detailed picture of these trends can be found in \Cref{tab:results}. Similar patterns are shown by the RMSE and the CRPS (see Table S1 and S2 and Fig. S1 in the supplementary information), where the ECMWF S2S model presents the lower RMSE values, while \textit{DiffScale} considerably improves the performance across all lead times. 

Across the metrics considered in this study, the \textit{DiffScale} configurations outperform the ECMWF S2S model and climatology benchmarks, particularly at earlier lead times, demonstrating their robustness and adaptability. At coarser resolutions (S/3 and S/4), \textit{DiffScale} retains its skill over longer lead times, highlighting its ability to balance fine-scale dynamics with spatial variability. In general, \textit{DiffScale} shows a strong performance across resolutions, but challenges remain at finer scales for longer lead times.

\begin{table}[t]
    \centering
    \setlength{\tabcolsep}{1.5pt}
    \scalebox{1}{
    \begin{subtable}{0.5\textwidth}
        \centering
        \scalebox{1}{
            \begin{tiny}
            \begin{tabular}{llcccccc}
                \toprule
                \multirow{2}{*}{} & \multirow{2}{*}{Method}      & \multicolumn{6}{c}{MAE $\downarrow$}\\
                            \cmidrule{3-8}
                            &  & Bin$_1$ & Bin$_2$ & Bin$_3$ & Bin$_4$ & Bin$_5$ & Bin$_6$\\
                \midrule
                \multirow{5}{*}{$S$} 
                & Climatology                 & 0.517 & 0.508 & 0.525 & 0.527 & 0.530 & 0.504 \\
                & ECMWF S2S                  & 0.300 &  0.408 &  0.502 & 0.538 & 0.562 & 0.527  \\
                    \cdashlinelr{2-8}
                & Disffscale (sf \& lr-df)          & 0.305 & 0.399 & 0.498 & 0.522 & 0.533 & 0.502 \\
                & Disffscale (lr-ws, sf \& lr-df)   & 0.301 & 0.390 & 0.492 & \bf{0.511} & 0.525 & \bf{0.494}  \\
                & Disffscale (lr-ws \& sf)     & \bf{0.260} & \bf{0.369} & \bf{0.490} & 0.523 & \bf{0.524} &  0.498 \\
                \midrule
                \multirow{5}{*}{$S/2$} 
                & Climatology                 & 0.517 & 0.509 & 0.526 & 0.527 & 0.531 & 0.505 \\
                & ECMWF S2S                     & 0.291 & 0.400 & 0.495 & 0.532 & 0.555 &  0.521 \\
                    \cdashlinelr{2-8}
                & Disffscale (sf \& lr-df)          & 0.305 &  0.404 & 0.475 &  0.510 &  0.580 & 0.496 \\
                & Disffscale (lr-ws, sf \& lr-df)      &  0.297 & 0.390 & 0.476 & \bf{0.500} & \bf{0.525} &  \bf{0.488} \\
                & Disffscale (lr-ws \& sf)           & \bf{0.255} & \bf{0.376} & \bf{0.470} & 0.507 & \bf{0.525} &  0.499 \\
                \midrule
                \multirow{5}{*}{$S/3$} 
                & Climatology                 & 0.525 & 0.517 & 0.533 & 0.534 & 0.539 & 0.512 \\
                & ECMWF S2S                      & 0.281 & 0.394 &  0.491 & 0.528 & 0.552 & 0.517 \\
                    \cdashlinelr{2-8}
                & Disffscale (sf \& lr-df)          &  0.297 & 0.390 & 0.490 & 0.508 & 0.517 & 0.489 \\
                & Disffscale (lr-ws, sf \& lr-df)   & 0.294 & 0.388 & 0.486 & 0.505 & 0.\bf{507} & \bf{0.492} \\
                & Disffscale (lr-ws \& sf)           & \bf{0.251} & \bf{0.367} & \bf{0.475} & \bf{0.504} &  0.510 & 0.501 \\
                \midrule
                \multirow{5}{*}{$S/4$}
                & Climatology                 & 0.519 & 0.511 & 0.527 & 0.528 & 0.532 & 0.507 \\
                & ECMWF S2S                      & 0.268 & 0.381 &  477 &  0.541 & 0.539 & 0.504  \\
                    \cdashlinelr{2-8}
                & Disffscale (sf \& lr-df)          &  0.293 & 0.389 & 0.479 & 0.495 & 0.507 & \bf{0.477} \\
                & Disffscale (lr-ws, sf \& lr-df)   & 0.287 & 0.382 & 0.481 & \bf{0.494} & 0.509 & \bf{0.477} \\
                & Disffscale (lr-ws \& sf)          & \bf{0.246} & \bf{0.362} & \bf{0.470} & \bf{0.494} & \bf{0.508} &  0.487 \\
                \bottomrule
            \end{tabular}
            \end{tiny}
        }    
          
        \caption{MAE scores. Reported are the best scores observed during training for respective bins.}
         \label{tab:MAE}
    \end{subtable}
    }
    \scalebox{1}{
    \begin{subtable}{0.5\textwidth}
       \centering
        \scalebox{1}{
            \begin{tiny}
                \begin{tabular}{llcccccc}
                    \toprule
                    \multirow{2}{*}{} & \multirow{2}{*}{Method}      & \multicolumn{6}{c}{CRPSS $\uparrow$}\\
                                \cmidrule{3-8}
                                &  & Bin$_1$ & Bin$_2$ & Bin$_3$ & Bin$_4$ & Bin$_5$ & Bin$_6$\\
                    \midrule
                    \multirow{4}{*}{$S$} 
                    & ECMWF S2S                    &  0.334 & 0.158 & 0.013 & -0.054 & -0.093 & -0.065  \\
                    \cdashlinelr{2-8}
                    & Disffscale (sf \& lr-df)          & 0.355 & 0.167 & 0.007 & -0.019 & -0.036 &  -0.12  \\
                    & Disffscale (lr-ws, sf \& lr-df)   &  0.360 & 0.189 & 0.028 & \bf{0.008} & \bf{-0.023} &  \bf{-0.001} \\
                    & Disffscale (lr-ws \& sf)          & \bf{0.467} & \bf{0.248} & \bf{0.040} & -0.010 & -0.015 & -0.005\\
                    \midrule
                    \multirow{4}{*}{$S/2$} 
                    & ECMWF S2S                      & 0.360 & 0.178 & 0.030 & -0.038 & -0.077 & -0.048 \\
                    \cdashlinelr{2-8}
                    & Disffscale (sf \& lr-df)          & 0.357 & 0.157 & 0.058 & 0.001 & -0.025 & -0.012 \\
                    & Disffscale (lr-ws, sf \& lr-df)   & 0.367 & 0.184 & 0.062 & \bf{0.022} & -0.016 & \bf{0.004}  \\
                    & Disffscale (lr-ws \& sf)          &  \bf{0.475} & \bf{0.234} & \bf{0.079} & 0.013 & \bf{-0.010} & -0.012 \\
                    \midrule
                    \multirow{4}{*}{$S/3$} 
                    & ECMWF S2S                      & 0.396 & 0.208 & 0.057 & -0.012 & -0.049 & -0.021\\
                    \cdashlinelr{2-8}
                    & Disffscale (sf \& lr-df)          &  0.385 & 0.204 & 0.046 & 0.027 & 0.010 & \bf{0.023} \\
                    & Disffscale (lr-ws, sf \& lr-df)   & 0.386 & 0.203 & 0.059 & 0.030 & \bf{0.031} & 0.019 \\
                    & Disffscale (lr-ws \& sf)          & \bf{0.492} & \bf{0.266} & \bf{0.089} & \bf{0.038} & 0.030 & 0.003\\
                    \midrule
                    \multirow{4}{*}{$S/4$} 
                    & ECMWF S2S                      &  0.423 & 0.230 & 0.074 & 0.003 & -0.037 & -0.005 \\
                    \cdashlinelr{2-8}
                    & Disffscale (sf \& lr-df)          &  0.389 & 0.190 & 0.058 & 0.044 & 0.020 & 0.043 \\
                    & Disffscale (lr-ws, sf \& lr-df)   & 0.395 & 0.208 & 0.054 & 0.042 & 0.016 & 0.042 \\
                    & Disffscale (lr-ws \& sf)          & \bf{0.500} & \bf{0.267} & \bf{0.084} & \bf{0.051} & \bf{0.023} & 0.025\\
                    \bottomrule
                \end{tabular}
            \end{tiny}
        }

        \caption{CRPSS scores. Reported are the best scores observed during training for respective bins.}
        \label{tab:CRPSS}
    \end{subtable}
    }
    \caption{Quantitative results for \textit{evaluation data}.}
    \label{tab:results}
\end{table}

\subsection{Qualitative assessment of spatial predictions}  \label{sec:eval-qual}

To further assess the model performance, we examine the spatial predictions from both the ECMWF S2S and \textit{DiffScale} models, focusing on their relative skill wind speed forecasts across the study region. All following qualitative assessments are made for the \textit{DiffScale} (lr-ws \& sf) experiment, which achieves the highest skill scores across most lead times.  Additional details on the spatial skill scores for the \textit{DiffScale} (lr-ws, sf \& lr-df) and \textit{DiffScale} (sf \& lr-df) experiments can be found in Figures S2–S5 and S6–S9 in the supplementary information. A broad overview of the generative behavior at the highest scaling factor and respective resolution is visualized in \Cref{fig:fig1}, where the dataset mean of ERA5 ECMWF and \textit{DiffScale} (lr-ws \& sf) is plotted. We observe a reconstruction of details, which will be highlighted in the following subsections. See S2 in the supplementary information for evaluations of additional \textit{DiffScale} configurations.

\paragraph{Spatial MAE and CRPS} We begin by assessing the spatial variability of skill scores displayed in \Cref{fig:fig_eval}. \Cref{fig:figmae} illustrates the spatial distribution of the MAE over the entire domain. In agreement with the quantitative results, \textit{DiffScale} shows better performance across lead times and scaling factors. The expected decline in skill scores with increasing lead time is evident in both the ECMWF S2S and \textit{DiffScale} models across most of the study region. However, the northern coastal regions show the most pronounced decrease in the predictive skill, with the lowest values shown by the ECMWF S2S model. Similarly, \Cref{fig:figcrps} illustrates the spatial variability of the CRPS. As with the deterministic MAE, CRPS values are lower at shorter lead times but increase significantly with lead time, particularly over coastal regions. Notably, \textit{DiffScale} shows a substantial improvement in CRPS values over the northern part of the domain compared to the deterministic MAE (see \Cref{fig:figmae}). This suggests that \textit{DiffScale} effectively generates probabilistic outputs that enhance the predictive skill of S2S forecasts.

This improvement in CRPS highlights the ability of \textit{DiffScale} to better capture the inherent uncertainty in long-range forecasts, particularly in spatial regions where ECMWF predictions degrade rapidly. The pronounced increase in MAE over coastal areas suggests that these regions exhibit greater forecast uncertainty, likely due to complex land-sea interactions that make accurate predictions challenging.

\paragraph{Spatial bias} Complementary to the spatial MAE and CRPS, \Cref{fig:figbias} shows the spatial bias (relative to the target ERA5) for the ECMWF S2S and \textit{DiffScale} calculated as the mean of the bias across all forecast times for each lead time and resolution. Overall, \textit{DiffScale} exhibits lower bias compared to the ECMWF S2S model, indicating improved skill in accurately predicting wind speed over the domain. The bias patterns shown by the ECMWF S2S are more pronounced, with systematic overestimation in coastal and offshore regions and underestimation in inland areas. This suggests that the ECMWF S2S model captures large-scale wind patterns but struggles to resolve fine-scale variations, which results in regionally consistent biases. Our model, on the other hand, leads to reduced biases across most regions, reflecting its ability to better capture localized ws10 variations. We observe this improvement  across all lead times, with \textit{DiffScale} exhibiting lower bias compared to the ECMWF S2S model, even as the lead time increases. However, we notice that while the \textit{DiffScale} method generally outperforms the ECMWF S2S model in reducing bias, for certain regions there are still notable discrepancies, particularly for longer lead times (\Cref{fig:figbias}). Moreover, at longer lead times, a more pronounced change in the sign of bias can be observed, which can be attributed to inconsistent atmospheric circulation patterns during the training and testing phases \citep{Bouallègue2024}. These inconsistencies might also be due to the small size of the test dataset, comprising only 104 forecast times (i.e., 104 initialization times for the testing data during 2021), which could limit the robustness of the results. Despite these limitations, the biases in \textit{DiffScale} predictions are generally lower and more geographically constrained than those observed in ECMWF S2S, providing accurate predictions of ws10 forecasts. 
\paragraph{Spatial Anomaly Correlation Coefficient} The spatial distribution of the Anomaly Correlation Coefficient (ACC) illustrates the variability of forecast skill over our region of study (\Cref{fig:figacc}). As anticipated, the highest ACC values are observed for shorter lead times, with a considerable decline as lead time increases. This trend is consistent for both the ECMWF S2S and \textit{DiffScale} models, as well as the different resolutions. However, \textit{DiffScale} consistently outperforms ECMWF S2S in terms of ACC values across all lead times. This performance gap is particularly evident in the southern regions, with more complex orography, such as the Alps. In locations, the ECMWF S2S forecasts exhibit ACC values near zero or negative, even for the shorter lead times, indicating limited skill in capturing ws10 conditions in this area.

Despite the overall improved forecast skill demonstrated by \textit{DiffScale}, a decrease in ACC values is observed in some southern regions starting from week 3 (i.e., $Bin_4$). This decline underscores the inherent challenges associated with predictability over longer time scales, particularly in areas characterized by complex topography. The observed reduction in skill suggests that while \textit{DiffScale} excels in providing accurate forecasts in the short term, its ability to sustain high skill over extended lead times, especially in such challenging terrains, is limited. This pattern further emphasizes the difficulty of achieving reliable long-range forecasts, particularly in regions with complex topography where atmospheric predictability is inherently limited. Still, this highlights the added value of incorporating advanced post-processing techniques to improve S2S forecast skill in regions where numerical models struggle.
These findings underline \textit{DiffScale}'s potential for high-resolution wind forecasting in applications such as renewable energy planning and extreme weather preparedness.

\section{Discussion}\label{conclusions}

This work introduces a novel approach for downscaling continuous spatial resolutions of S2S forecasts of surface wind speed across multiple lead times. Building on the capabilities of diffusion models, \textit{DiffScale} continuously resolves finer wind speed resolutions across varying time scales. 
We demonstrate the effectiveness of \textit{DiffScale} for post-processing S2S forecasts, a domain in which the adoption of deep learning-based methods is still limited.
For our experimental framework, we use four scaling factors and select several predictor configurations, including both atmospheric and static fields from the S2S database \citep{Vitart2017}. Our method aimed not only to correct the bias inherent in the numerical model, but also at enhancing the spatial resolution , using ERA5 reanalysis as ground truth. \textit{DiffScale} was trained using perturbed ensemble members from reforecasts of the ECMWF S2S model and tested on real forecasts of ECMWF S2S. While only 10 ensemble members per input were generated during the testing phase, it is important to highlight that \textit{DiffScale} allows for a variable number of ensemble predictions. 

The results demonstrated the ability of \textit{DiffScale} to post-process S2S forecasts of surface wind speed, outperforming the ECMWF S2S and the climatology used as a benchmark. The quantitative assessment  showed that \textit{DiffScale} consistently achieved higher skill scores across all lead times, with an expected decline in performance at longer lead times (e.g., Bins 5 and 6). Overall, performance differences among \textit{DiffScale} experiments were minimal, with the configuration using only static fields and low forecast resolution as inputs showing slight improvements at shorter lead times. At longer lead times, additional atmospheric inputs appeared to contribute to performance. However, further investigation is needed to quantify their impact on the model’s predictive skill over longer time scales.

The qualitative evaluation highlighted regional variations in predictive skill scores, with \textit{DiffScale} demonstrating significantly improved performance compared to the ECMWF S2S model. The ECMWF S2S exhibited higher biases, particularly over northern coastal areas and the southern domain characterized by complex orography. Among the metrics used for the qualitative assessment, we observe a notable improvement in the spatial CRPS, especially in those regions where deterministic metrics, such as MAE, show a greater increase with lead time. This suggests that \textit{DiffScale} effectively captures uncertainty at longer lead times.
While the methodological framework presented here illustrates the potential of diffusion models for super-resolution and ensemble generation, we acknowledge computational constraints that limited both the scope of our experiments and our choice to focus on a relatively small domain in central Europe. It is also worth noting that the limited training data, consisting of only 20 years of reforecasts, may have influenced performance, potentially resulting in poorer performance than with a larger time horizon. 

Despite these potential limitations, \textit{DiffScale} presents a flexible framework for effectively representing wind speed variability at different spatial scales, while enhancing the predictive skill of numerical forecasts. These strengths make \textit{DiffScale} a valuable tool for S2S forecasting applications, supporting decision-making in sectors such as energy and resource management.

\section{Method}\label{sec11}

In the following, we outline the methods used to collect data and conditioning inputs, as well as the working principles of the proposed \textit{DiffScale} model.

\begin{table}[tb]
    \centering
    \begin{footnotesize}
        \begin{tabularx}{\linewidth}{p{2.8cm}XXX}
        \toprule
        \textbf{Long Name}                   & \textbf{Short Name}   & \textbf{Pressure levels (hPa)} & \textbf{Res. (\(^{\circ}\))} \\ \midrule
        temperature (at 2m)            & t2m                  & -                          & 1.25   \\
        Mean sea level pressure              & mslp                 & -                          & 1.25   \\
        Zonal wind                           & u                    & 300, 925                   & 1.25   \\
        Meridional wind                      & v                    & 300, 925                   & 1.25   \\
        Geopotential                         & Z                    & 500                        & 1.25  \\
        10m wind speed                       & ws10                 & -                          & 1.25   \\
        Land-sea mask                        & -             & -                          & 0.25   \\ 
        Geopotential (surface)                            & Orography            & -                          & 0.25   \\ 
        \bottomrule
        \end{tabularx}
    \end{footnotesize}
\caption{Summary of input variables.}
\label{tab:input_variables}
\end{table}

\input{figure_model_wide}
\subsection{Data}\label{sec:data}

The S2S database is an extensive database for the subseasonal time scale that emerged from a global research initiative that aims at bridging the gap between short- and medium-range forecasts (up to two weeks) and seasonal predictions \citep{Vitart2017}. The forecasts and retrospective forecasts (reforecast) data are derived from the European Centre for Medium-Range Weather Forecasts (ECMWF), accessible via the Subseasonal-to-Seasonal (S2S) Prediction Project Database \cite{Vitart2017}. ECMWF generates reforecasts simultaneously with the real-time forecasts, which are computed twice a week (Mondays and Thursdays). They consist of 11 ensemble members (10 perturbed and 1 control), covering a forecast lead time up to 46 days with a spatial resolution of 1.5\si{\degree}. The reforecasts data spans from 2001 to 2020. In this study, only the perturbed members of both forecasts and reforecasts are utilized. The forecasts are the operational ensemble predictions consisting of 51 members (50 perturbed and 1 control). Reforecasts from 2001 to 2018 are used for training, while those in 2019 and 2020 are used for validation. For testing, we used the operational forecasts for 2021, which include model cycles Cy47r1, Cy47r2, and Cy47r3, switching in May and then October of 2021, respectively. As noted in \cite{Bouallègue2024}, the model cycle Cy47r2 includes an improvement in the vertical levels, which is not considered in the training data. The operational forecasts, used as test data, include a total of 104 forecast times (i.e., twice per week initialization times). It is worth noting that the evaluation was conducted using 10 ensemble members, consistent with the number of members used during training. However, ensemble size is not expected to significantly impact results. All S2S fields come at a resolution of 1.5\si{\degree}.

Atmospheric variables, namely 2m surface temperature, mean sea level pressure, zonal wind (u), meridional wind (v) at 300, 925 hPa, geopotential at 500 hPa and 10m wind speed are used as input data. Additionally, high-resolution static variables, including land-sea mask and orography data are used as inputs. \Cref{tab:input_variables} summarizes the additional input variables. 

To enhance both the skill and resolution of S2S forecasts, the 10m wind speed data from ERA5 reanalysis \cite{Hersbach2020} at its original spatial resolution (0.25\si{\degree}) is used as the high-resolution target dataset.

\subsection{Diffusion models for continuous spatiotemporal downscaling} \label{sec:back}

\paragraph{Background} Since the inception of diffusion models \citep{sohl2015deep, ho2020denoising}, the formulation of the diffusion process itself has undergone rethinking. Rather than constraining the process to discrete steps in a Markov chain, we leverage a continuous-time perspective via an SDE \citep{song2020score}. In the following, we give a brief overview of the continuous-time formulation of score-based diffusion models through SDEs. We point out that driftless diffusion processes can be entirely described by properties of the variance schedule, as shown in \citet{Karras2022elucidating}.

We first provide a brief overview of the continuous-time formulation of score-based diffusion models through the lens of SDEs. 
\citet{song2020score} propose modeling the distribution transformation process from data to noise as a stochastic process in continuous time, following the forward dynamics
\begin{align}
    \dd \x &= \f(t) \x \dt + \g(t) \dw, \label{eq:score:forw}
\end{align}
where $\w$ is a standard Wiener process with a continuous drift function $\f:[0,T]\to\mathbb{R}$ and continuous diffusion function $\g:[0,T]\to\mathbb{R}$. 
Remarkably, this formulation provides an exact reverse-time process with dynamics \citep{anderson1982}
\begin{align}
    \dd \x &= \left[ \f(t) \x - \g(t)^2 \nabla_\x \log p_t(\x)\right] \ \dt + g(t) \ \dw, \label{eq:score:rev}
\end{align}
where $\nabla_\x \log p_t$ is the unknown score function associated with the marginal density $p_t$ of $\mathbf{x}$ at time $t\in[0,T]$.
The unknown score function is approximated using a parameterized score model $\scoreF$, which is trained via score-matching \cite{song2020score}.

To efficiently sample from the forward process at any time, we can adopt a probabilistic perspective. Conditioning the forward process on its starting value $x(0)$ yields the closed-form transition kernel \citep{song2020score, Karras2022elucidating}
\begin{align}
    \pkernel{x(t)}{x(0)}    & = \N{x(t);\ s(t) x(0), s(t)^2 \sigma(t)^2\eye},
\end{align}
where transitions, related to the function $\f$ are described by $s(t)  = \exp\left(\int_0^t f(t)\right)$ and the dynamic relation of the function $\f$ and the diffusion function $\g$ is incorporated into the variance schedule $\sigma(t)$. Furthermore $g$ can be expressed as $g(t) = s(t) \sqrt{2 \sigma(t) \frac{\partial\sigma}{\partial t}(t)}$ \citep{Karras2022elucidating}. Note that a driftless forward process can be parameterized by setting $f \equiv 0$, inducing $s \equiv 1$, which results in the process being defined by $\sigma(t)$. In the following we will refer to the definition of the variance schedule $\sigma(t)$ to define the diffusion process used in this work.

\paragraph{DiffScale}
Using a diffusion model, we aim to perform continuous spatial upsampling and lead time interpolation within a single model. We acknowledge that forecasts with higher lead times tend to be less precise and should be treated differently to more precise predictions with lower lead times when upsampling (see \Cref{fig:numeric-MAE}). Given a forecast with low spatial resolution $\x \in \mathbb{R}^{H \times W}$ at lead time $l \in \mathbb{R}_{\geq 1}$, we assume that a true probability density
$p_{\alpha}(\x^{(\alpha)} | \x, \tau)$
exists, where $\tau = \{\alpha, l\}$ and $\x^{(\alpha)} \in \mathbb{R}^{\alpha H \times \alpha W}$ is a higher resolution forecast, upsampled by a factor $\alpha \in \mathbb{R}_{\geq 1}$ from $\x$.

\input{binning}

Drawing from the unknown density $p_{\alpha}$ would yield an optimal method for upsampling. We approximate drawing $\hat{\x}^{(\alpha)} \sim p_{\alpha}(\x^{(\alpha)} | \x, \tau)$ by drawing from the reverse diffusion process with our score based diffusion model, trained on forecasts at different spatial resolutions and lead times. We can achieve this by using an ODE or SDE solver method for the reverse process and conditioning the diffusion model on the low resolution input $\x$, as well as the scaling factor $\alpha$ and lead time $l$. A remarkable benefit of this conditional probabilistic perspective is that we can produce samples for continuous lead time $l$ and scaling factor $\alpha$ without relying on autoregression or any other form of discretization for the temporal axis (w.r.t. lead time). The model learns to produce samples at any scale and lead time by matching the density $p_{\alpha}$ directly, rather than propagation through lead time. We want to emphasize the distinction between $p_{\alpha}$ and the data distribution of $x^{(\alpha)}(0)$. The density $p_{\alpha}$ induces the distribution of sampling from the optimal predictor, given the provided context, whereas $x^{(\alpha)}(0)$ is the true, unconditional data distribution. We aim to find specific samples of  $x^{(\alpha)}(0)$ that fit a set of conditional inputs $\tau$ via $p_{\alpha}$. Depending on the architecture, the model can produce more or less finely pixelated images w.r.t. the upsampling factor at a fixed output size, which is equivalent to a feasible discretization at a variable output size, but can be scaled arbitrarily large.

Furthermore, we acknowledge that in the absence of strongly constraining context the density $p_{\alpha}(\x^{(\alpha)} | \x, \tau)$ is likely to exhibit high variance. To mitigate the challenge of producing high fidelity samples in a probabilistic framework, we also experimented with conditioning the model on additional weather variables and regional priors $R$ listed in \Cref{tab:input_variables}, resulting in $\tau = \{R, \alpha, l\}$.

Using the proposed \textit{DiffScale} method we combine both upsampling and forecast correction (i.e., prediction) for continuous spatial resolution and lead time within one model. An additional advantage of our approach is that it does not rely on time-series information, rather it leverages information from the numerical prediction and regional priors. This approach enhances the sample-efficiency and the applicability of the model without any dependence on 3D-convolutions or auto-regression.

\subsection{Hyperparameters}

\paragraph{Variance schedule}

There have been many efforts to optimize the definition of the forward process with respect to the fidelity of the resulting samples \citep{ho2020denoising, song2020score, Karras2022elucidating}. A well-established definition of the forward process that has withstood the test of time is the variance exploding (VE) variance schedule
\begin{align}
    \sigma(t) & = \sigma_{\min} \left(\frac{\sigma_{\max}}{\sigma_{\min}}\right)^t,
\end{align}
where $t \in [0,1]$. The forward process associated with the VE variance schedule is a driftless process with $f\equiv0$. We follow \citet{song2020score} and set $\sigma_{\max}=50$ and $\sigma_{\min}=0.01$.

\begin{figure*}[tb]
    \centering
    \includegraphics[width=0.9\linewidth]{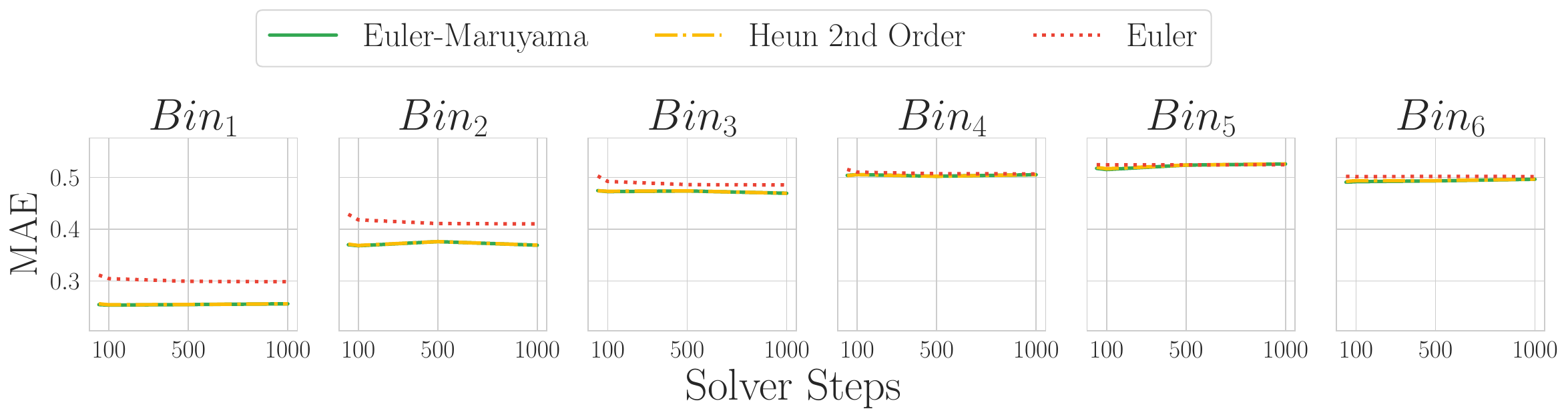}
    \caption{Evaluation of the MAE for different solver methods and number of function evaluations (solver steps) per sampling of downscaled $\hat{x}^{(\alpha)}$. Displayed is the mean over all scaling factors per bin and number of solver steps.
    }
    \label{fig:ablation}
\end{figure*}

\paragraph{SDE solver method}

Resolving the true density $p_{\alpha}$ is computationally infeasible. However, leveraging the working principle of diffusion models, we can obtain an approximation of $p_{\alpha}$. To obtain an approximation of sampling $\hat{\x}^{(\alpha)} \sim p_{\alpha}(\x^{(\alpha)} | \x, \tau)$, we need to solve the reverse diffusion process with a conditional score function. We evaluated three common solver methods at varying numbers of equidistant steps. Results showed that while the probability flow solver (Euler method) \citep{song2020score} did not yield good results, the 2nd Order Heun Solver \citep{Karras2022elucidating} demonstrated reasonable performance as an ODE solver (at the cost of an additional model evaluation per step). Neither method outperformed the Euler-Maruyama method \citep{song2020score}, which solves the underlying SDE of the reverse process, as shown in \Cref{fig:ablation}. We also evaluated the number of discretization steps required for our approach in \Cref{fig:ablation}, with a selection of 50, 100, 500, and 1000 solver steps. We observed that near-optimal performance was already reached at 50 to 100 discretization steps for all solver methods respectively, without any observable decrease in performance w.r.t. MAE after 100 steps. The Heun Solver performs identically to the Euler-Maruyama solver -- however each solver step of the Heun solver requires two model evaluations, making it more costly than the Euler-Maruyama method. All results are reported for the Euler-Maruyama method with 100 discretization steps of the reverse process, unless stated otherwise.

\subsection{Evaluation}

\paragraph{Evaluation metrics}
The quality of the \textit{DiffScale} is assessed through several well-established deterministic metrics, including the bias to evaluate systematic errors, the root mean square error (RMSE) and mean squared error (MSE) to measure the overall model accuracy and the magnitude of errors, respectively. Additionally, we use the Anomaly Correlation Coefficient (ACC), a widely used metric in the verification of spatial fields,  to assess the model's ability to represent observed anomalies. 

For probabilistic evaluation, the continuous ranked probability score (CRPS) is used to assess how well the forecast distributions align with observed outcomes, reflecting both forecast reliability and sharpness \cite{Hersbach2020}. Then, the continuous ranked probability skill score (CRPSS) is used to assess whether the forecasts improve or degrade relative to a reference forecast, in this case, the climatology.

We compare \textit{DiffScale} with the climatological forecasts as a well-established benchmark, commonly used in S2S forecasting to provide a competitive point of reference \citep{Horat2024, Mouatadid2023}. In addition, S2S forecasts, bilinearly interpolated to each target resolution without applying any bias correction, also serve as a baseline for comparison.

All these metrics are calculated for both the validation and test datasets to ensure consistency and robustness across different subsets of the data. However, unless otherwise specified, the results discussed in the following sections are based on the test dataset (i.e., real-forecast data). 

\paragraph{Evaluation of lead time}
While \textit{DiffScale} is agnostic to the lead time and is designed to generate predictions at all lead times,  we adopt a lead time binning approach to facilitate targeted evaluation without considering all lead times individually. This method divides the $46$ days lead time range into smaller subsets (''bins``) to focus on specific temporal intervals. Initially, our evaluation strategy sampled one lead time per week. This weekly binning approach ensured coverage of one lead time per week across the 46 days. However, preliminary results revealed a considerable variability in model performance within the first week of lead times, which suggested that refining our binning strategy by introducing smaller bins for the first 15 lead times would provide a better understanding of model performance.
This refined binning strategy allowed us to focus on critical variations in short-term forecast skill while maintaining an overview of performance over the full forecast horizon. See \Cref{fig:binning} for a visualization of the mapping.

\clearpage \newpage

\begin{figure*}[h]
    \centering
    \includegraphics[width=0.925\textwidth]{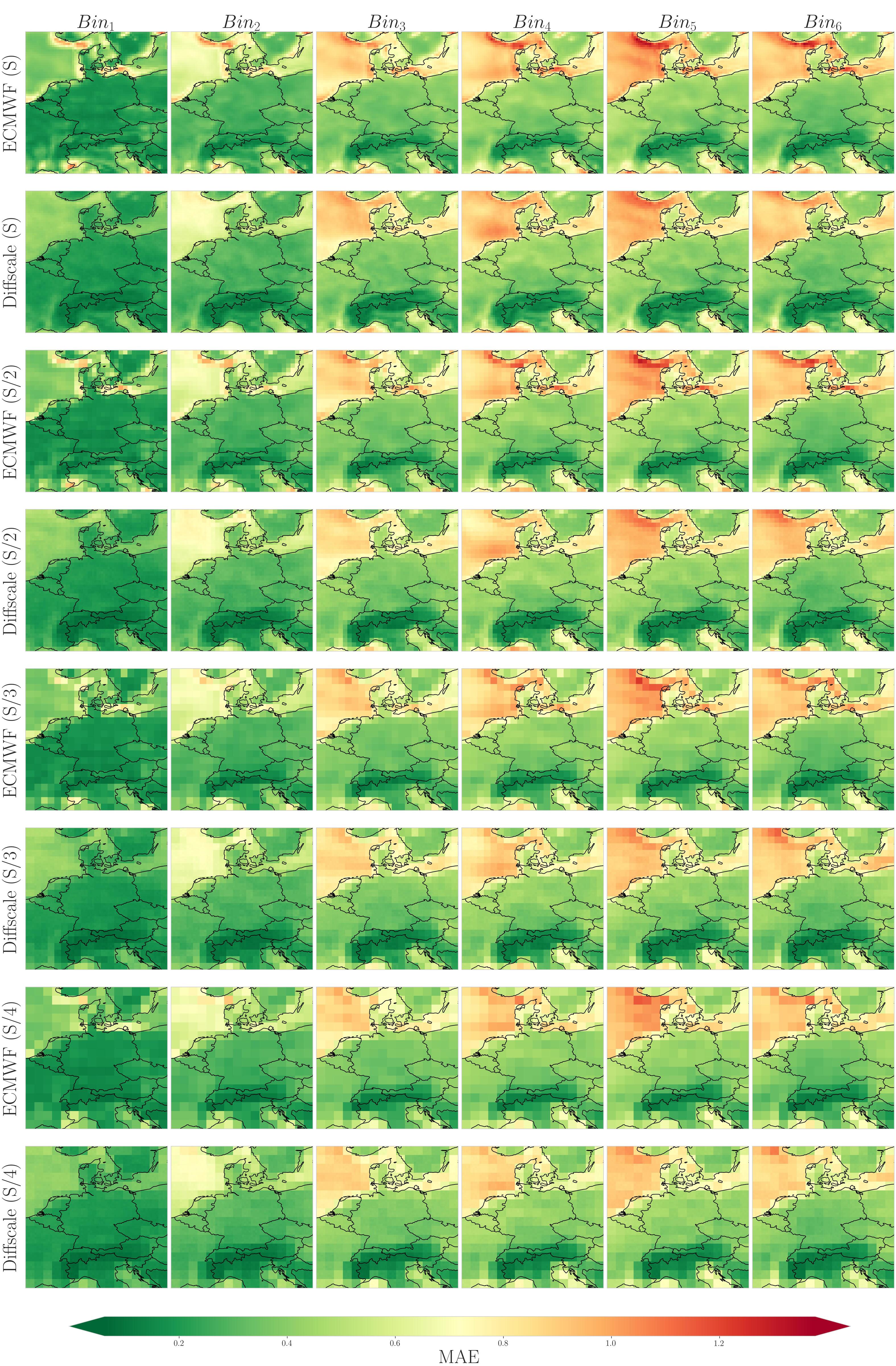}
    \caption{Spatial distribution of the MAE for the ECMWF and the \textit{DiffScale} (lr-ws \& sf) model for each resolution and lead time bins.}
    \label{fig:figmae}
\end{figure*}

\begin{figure*}[h]
    \centering
    \includegraphics[width=0.925\textwidth]{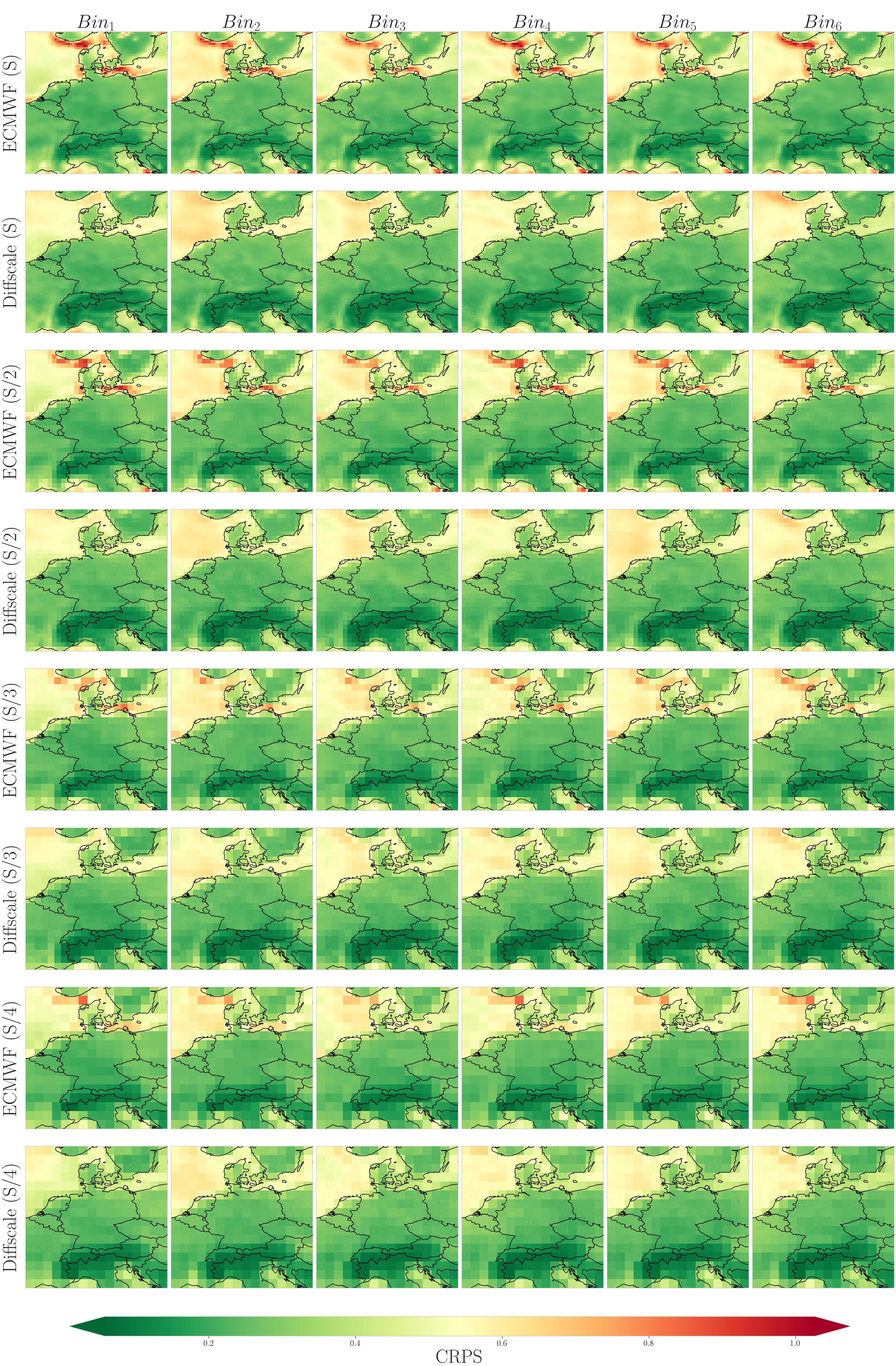}
    \caption{Spatial distribution of the CRPS for the ECMWF and the \textit{DiffScale} (lr-ws \&  sf) model for each resolution and lead time bins.}
    \label{fig:figcrps}
\end{figure*}

\begin{figure*}[h]
    \centering
    \includegraphics[width=0.925\textwidth]{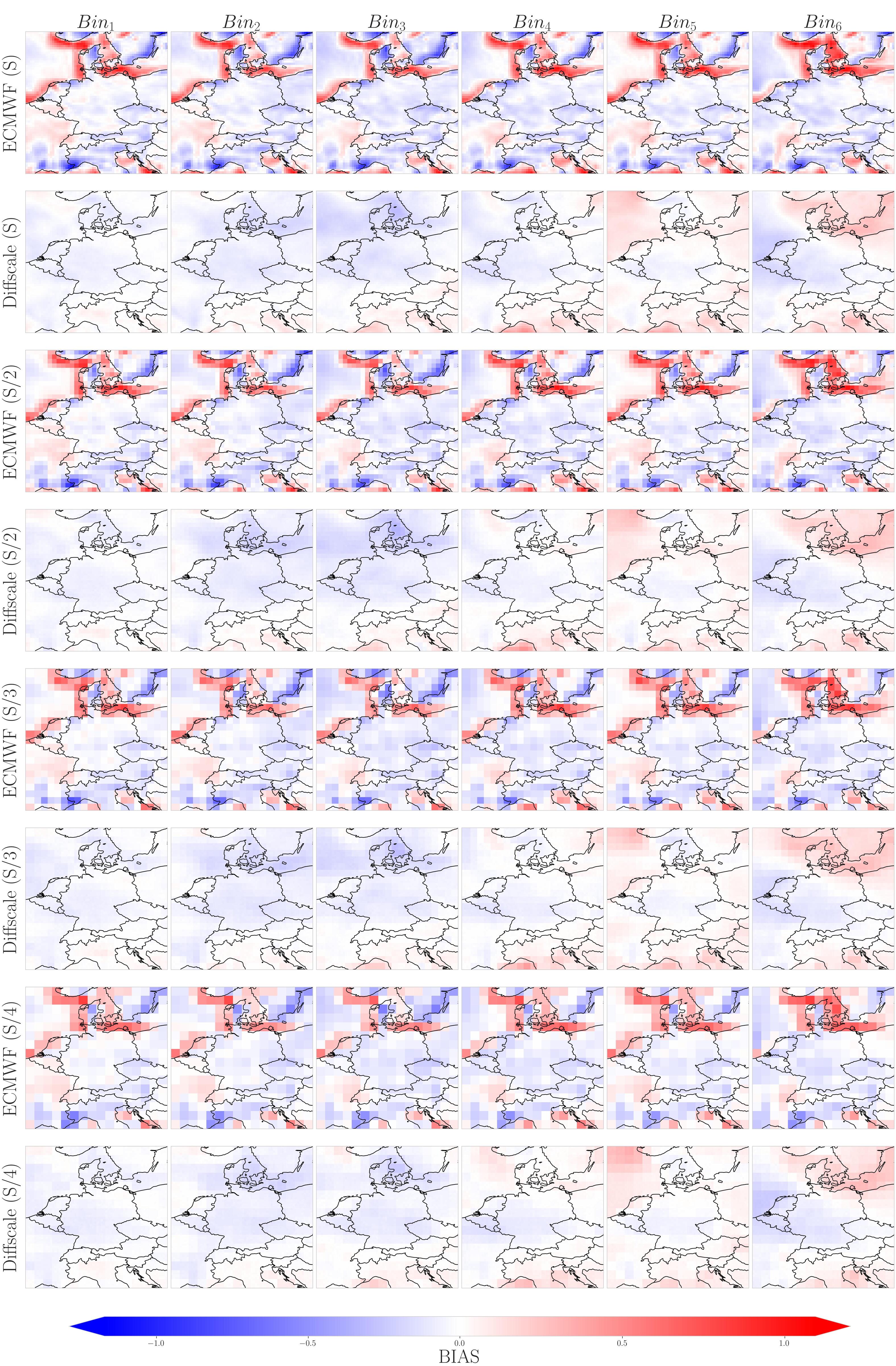}
    \caption{Bias of ws10m (m/s) calculated at each spatial resolution and lead time bin for the ECMWF and \textit{DiffScale}(lr-ws \& sf) model.}
    \label{fig:figbias}
\end{figure*}

\begin{figure*}[h]
    \centering
    \includegraphics[width=0.925\textwidth]{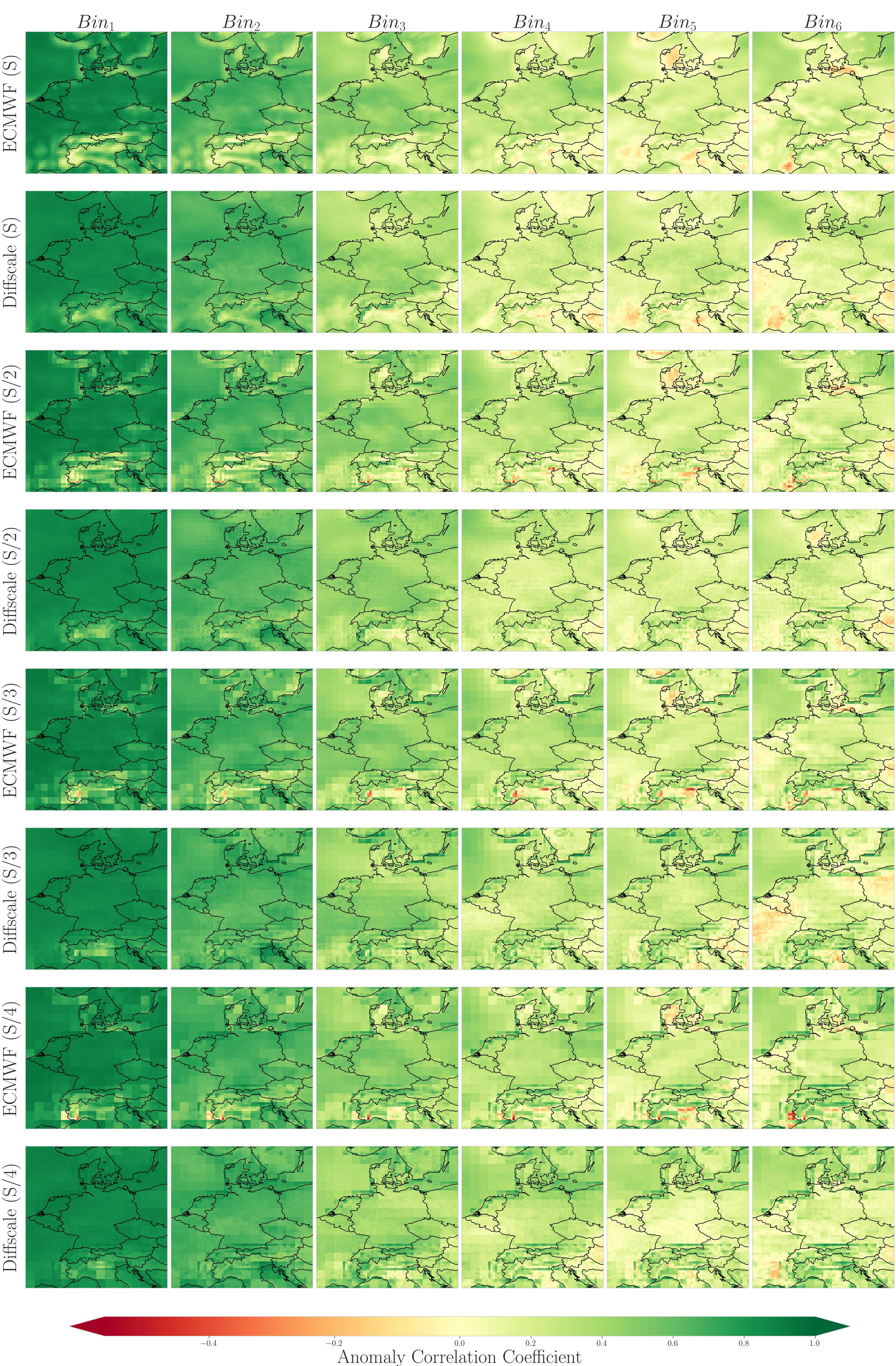}
    \caption{Spatial distribution of the anomaly correlation coefficient (ACC) for the ECMWF and the \textit{DiffScale}(lr-ws \& sf) model for each resolution and lead time bins. Note that the ACC is calculated considering all forecast times. }
    \label{fig:figacc}
\end{figure*}

\clearpage \newpage

\onecolumn
\section*{Declarations}

\subsection*{Supplementary information}
Supplementary information is provided in the appendix.
The same UNet architecture with attention in the bottleneck was chosen for all \textit{DiffScale} configurations and respective results reported in this work. All experiments were computed on A100 GPUs. All \textit{DiffScale} trainings converged within three A100 GPU days, respectively for all configurations with negligible differences in compute-time for different conditional inputs arising from model configurations (see \Cref{tab:experiments}).

\subsection*{Data availability} 
The data sets and code supporting this study are accessible to the public. The ERA5 data can be obtained from the Climate Data Store\footnote{\url{https://cds.climate.copernicus.eu/}}, and the S2S data is publicly accessible from ECMWF\footnote{\url{https://apps.ecmwf.int/datasets/data/s2s-realtime-instantaneous-accum-ecmf/levtype=sfc/type=cf/}}. All code is publicly accessible\footnote{\url{https://gitlab.hhi.fraunhofer.de/ai-aml/DiffScale}} to ensure the reproducibility of results.

\subsection*{Acknowledgements}

This work was supported by the European Union’s Horizon Europe research and innovation programme (EU Horizon Europe) as grant MedEWSa (101121192).

\subsection*{Author contributions} 

M.S. and N.O. have contributed equally to this work. M.S., N.O., and J.M. conceived the idea. M.S., N.O., and Y.X. conducted the experiments. M.S., N.O., and J.M. drafted the article and wrote the paper. All authors reviewed and revised the manuscript.

\subsection*{Competing Interests} 
The authors declare no competing interests.

\twocolumn

\clearpage\newpage
\onecolumn
\bibliographystyle{unsrtnat}
\bibliography{main}
\twocolumn
\onecolumn
\input{appendix}
\twocolumn

\end{document}

%% file: figure_model_wide.tex
\begin{figure*}[t]
    \centering
    \tikzset{
        ncbar angle/.initial=90,
        ncbar/.style={
            to path=(\tikztostart)
            -- ($(\tikztostart)!#1!\pgfkeysvalueof{/tikz/ncbar angle}:(\tikztotarget)$)
            -- ($(\tikztotarget)!($(\tikztostart)!#1!\pgfkeysvalueof{/tikz/ncbar angle}:(\tikztotarget)$)!\pgfkeysvalueof{/tikz/ncbar angle}:(\tikztostart)$)
            -- (\tikztotarget)
        },
        ncbar/.default=0.5cm,
    }
    
    \tikzset{round left paren/.style={ncbar=0.5cm,out=100,in=-100}}
    \tikzset{round right paren/.style={ncbar=0.5cm,out=80,in=-80}}
    
    \centering
    \scalebox{0.75}{
    \begin{tikzpicture}[scale=0.5]
        \draw[fill=cyan!15,draw=cyan!35,rounded corners] (-2.5,-2.25) rectangle ++ (22,-7.5);
        \draw[fill=orange!15,draw=orange!35,rounded corners] (-9,3.25) rectangle ++ (4,-5);
        \node[anchor=center] (x0) at (-7,0) {\includegraphics[width=1.5cm]{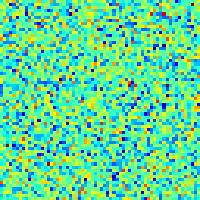}};
        \draw[fill=orange!15,draw=orange!35,rounded corners] (2.5,3.25) rectangle ++ (4,-5);
        \node[anchor=south] (x0label) at (x0.north) {$\hat{\x}^{(\alpha)}_T$};
        \node[anchor=center] (x1) at (4.5,0) {\includegraphics[width=1.5cm]{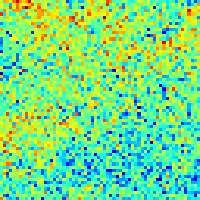}};
        \node[anchor=south] (x1label) at (x1.north) {$\hat{\x}^{(\alpha)}_{T-1}$};
        \node[anchor=center] (xi) at (15.25,0) {$\cdots$};
        \draw[fill=orange!15,draw=orange!35,rounded corners] (24,3.25) rectangle ++ (4,-5);
        \node[anchor=center] (xT) at (26,0) {\includegraphics[width=1.5cm]{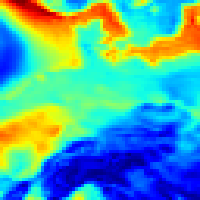}};
        \node[anchor=south] (xTlabel) at (xT.north) {$\hat{\x}^{(\alpha)}_0$};
        \node[anchor=west] (clabel) at (2, -3.5) {};
        \node[anchor=east] (c1) at (4.8, -4) {\includegraphics[width=1cm]{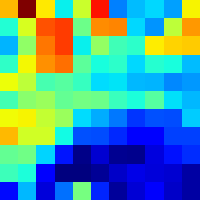}};
        \node[anchor=west, text width=7cm] (c1label) at (7, -4) {low resolution ECMWF mean $\x$};
        \node[anchor=east, opacity=.5] (c2) at (2.8, -6.2) {\includegraphics[width=1cm]{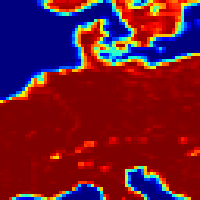}};
        \node[anchor=east, opacity=1] (c2) at (3, -6.3) {\includegraphics[width=1cm]{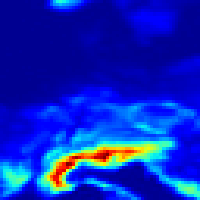}};
        \node[anchor=east, opacity=.5] (c2) at (5.2, -6.2) {\includegraphics[width=1cm]{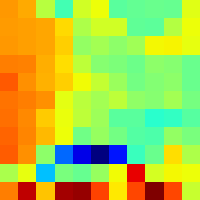}};
        \node[anchor=east, opacity=.5] (c2) at (5.4, -6.3) {\includegraphics[width=1cm]{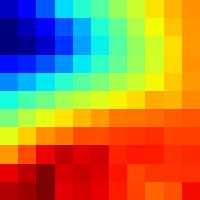}};
        \node[anchor=east, opacity=.5] (c2) at (5.6, -6.4) {\includegraphics[width=1cm]{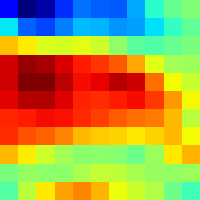}};
        \node[anchor=east, opacity=.5] (c2) at (5.8, -6.5) {\includegraphics[width=1cm]{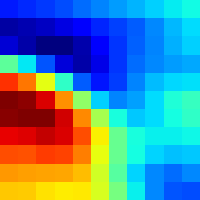}};
        \node[anchor=east, opacity=.5] (c2) at (6, -6.6) {\includegraphics[width=1cm]{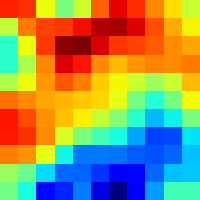}};
        \node[anchor=east, opacity=.5] (c2) at (6.2, -6.7) {\includegraphics[width=1cm]{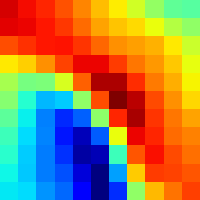}};
        \node[anchor=east, opacity=.5] (c2) at (6.4, -6.8) {\includegraphics[width=1cm]{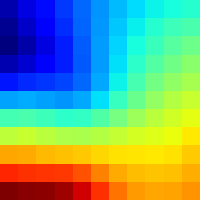}};
        \node[anchor=east, opacity=.5] (c2) at (6.6, -6.9) {\includegraphics[width=1cm]{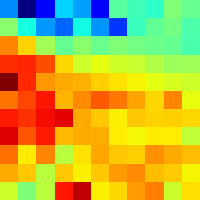}};
        \node[anchor=east, opacity=1.0] (c2) at (6.8, -7) {\includegraphics[width=1cm]{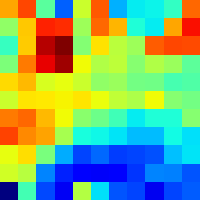}};
        \node[anchor=west, text width=6cm] (c2label) at (7, -6.8) {regional priors \& weather $R$};
        \node[anchor=east] (c3) at (4.8, -9) {\textcolor{black}{$\{\alpha, l\}$}};
        \node[anchor=west] (c3label) at (7, -9) {scaling factor \& lead time};
        \node[anchor=center] (emb) at (-1.75, -6) {$\tau = $};
        \node[inner sep=0] (p1) at (-2.75, -9){};
        \node[inner sep=0] (p2) at (-2.75, -2){};
        \node[inner sep=0] (p3) at (18, -2){};
        \node[inner sep=0] (px0) at (3, -2){};
        \node[inner sep=0] (px1) at (9, -2){};
        \node[inner sep=0] (pxi) at (14, -2){};
        \draw[thick, decorate, decoration={calligraphic brace,amplitude=15}](0,-9.5) -- (0,-2.5);
        \draw[thick, decorate, decoration={calligraphic brace,amplitude=15}](18,-2.5) -- (18,-9.5);
        \draw[-latex, thick] (x0) -- (x1);
        \draw[-latex, thick] (x1) -- (xi);
        \draw[-latex, thick] (xi) -- (xT);
        \draw[fill=gray!15, draw=gray!35, rounded corners] (-3.5,1.25) rectangle ++ (3.5,-1.75);
        \draw[fill=gray!15, draw=gray!35, rounded corners] (7.75,1.25) rectangle ++ (5,-1.75);
        \draw[fill=gray!15, draw=gray!35, rounded corners] (18.25,1.25) rectangle ++ (3.45,-1.75);
        \node[anchor=center] (sT) at (-1.75,-0){\tiny$\scoreF(\textcolor{orange}{\hat{\x}^{(\alpha)}_T}, t_T, \textcolor{cyan!100}{\tau})$};
        \node[anchor=center] at ($(sT) + (0,0.75)$) {\tiny solver step for};
        \node[anchor=center] (sT-1) at (10.25,-0){\tiny$\scoreF(\textcolor{orange}{\hat{\x}^{(\alpha)}_{T-1}}, t_{T-1}, \textcolor{cyan!100}{\tau})$};
        \node[anchor=center] at ($(sT-1) + (0,0.75)$) {\tiny solver step for };
        \node[anchor=center, rounded corners] (s1) at (20,-0){\tiny$\scoreF(\textcolor{orange}{\hat{\x}^{(\alpha)}_1}, t_1, \textcolor{cyan!100}{\tau})$};
        \node[anchor=center] at ($(s1) + (0,0.75)$) {\tiny solver step for };
        \draw[-, thick] (emb) -- ($(emb) + (0,4)$) -| ($(sT)-(0,0.5)$);
        \draw[-, thick] (emb) -- ($(emb) + (0,4)$) -| ($(sT-1)-(0,0.5)$);
        \draw[-, thick] (emb) -- ($(emb) + (0,4)$) -| ($(s1)-(0,0.5)$);
    \end{tikzpicture}
    }
        \caption{The \textit{DiffScale} Diffusion model with classifier free guidance $\tau$. Arrows symbolize numerical solver steps for the respective score, solving the reverse process to generate $\hat{\x}^{(\alpha)}$ given $\x$.}
    \label{fig:model}
\end{figure*}

%% file: binning.tex
\begin{figure*}[t]
    \begin{subfigure}[t]{\textwidth}
        \centering
        \includegraphics[width=0.7\textwidth]{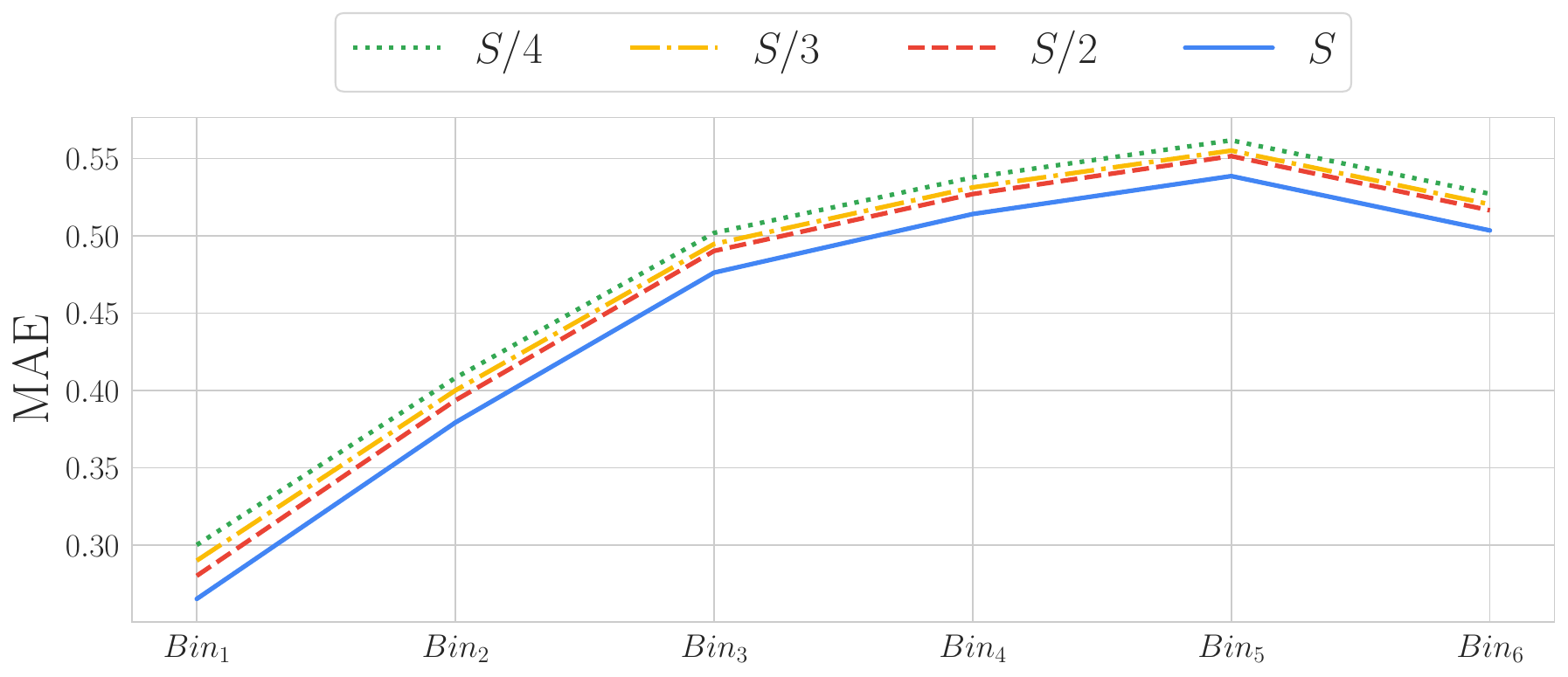}
        \caption{MAE of numeric ECMWF predictions over time; rescaled to target height via bilinear interpolation.}
        \label{fig:numeric-MAE}
    \end{subfigure}
    \begin{subfigure}[t]{\textwidth}
        \centering
        \begin{tikzpicture}
            \draw[thick, -latex] (-2 * 0.25, 0) -- (0.25*48 cm, 0);
            \foreach \i in {0, 7, 14, 21, 28, 35, 42}{
                \draw[thick] (0.25*\i cm,1mm) -- (0.25*\i cm,-1mm);
                \node[anchor=north] at (0.25*\i cm,-2mm) {\i};
            }
            \node[anchor=north] at (0.25*23, -6.75mm) {lead time [days]};
            \foreach \i/\j in {1/1, 4/2, 7/3, 10/4, 13/5}{
                \draw[fill=green!25,draw=black!85] (0.25*\i cm, 0.5mm) rectangle ++(0.25*3 cm,-1mm);
                \draw[decorate, decoration={calligraphic brace,amplitude=5}](0.25*\i cm, 1mm) -- (0.25*3cm + 0.25*\i cm, 1mm);
                \node[anchor=south] at (0.25*1.5cm + 0.25*\i cm, 2mm) {Bin$_{\j}$};
            }
            \draw[fill=blue!25,draw=black!85] (0.25*16 cm, 0.5mm) rectangle ++(0.25*30 cm,-1mm);
            \draw[decorate, decoration={calligraphic brace,amplitude=10}](0.25*16cm, 1mm) -- (0.25*46cm, 1mm);
            \node[anchor=south] at (0.25*15cm + 0.25*16 cm, 4mm) {Bin$_6$};
        \end{tikzpicture}
        \caption{Mapping of lead times to bins, for evaluation purposes. Short-range intervals are indicated in green and the long-range interval is indicated in blue.}
        \label{fig:binning} 
    \end{subfigure}
    \caption{Accumulation of errors for ECMWF numeric predictions over lead time and respective bins.}
\end{figure*}

%% file: appendix.tex
\appendix
\section{Appendix}

\paragraph{Supplementary content} The following contains supplementary information for the paper \textit{``DiffScale: Continuous Downscaling and Bias Correction of Subseasonal Wind Speed Forecasts using Diffusion Models''}. Additional tables and figures for the quantitative and qualitative assessment of the work described in the main text are provided below. We refer readers to the main text for a detailed description of our work.

\section{Quantitative assessment} \label{sup:quantitative}

\begin{figure*}[htbp]
    \centering
    \includegraphics[width=0.9\textwidth]{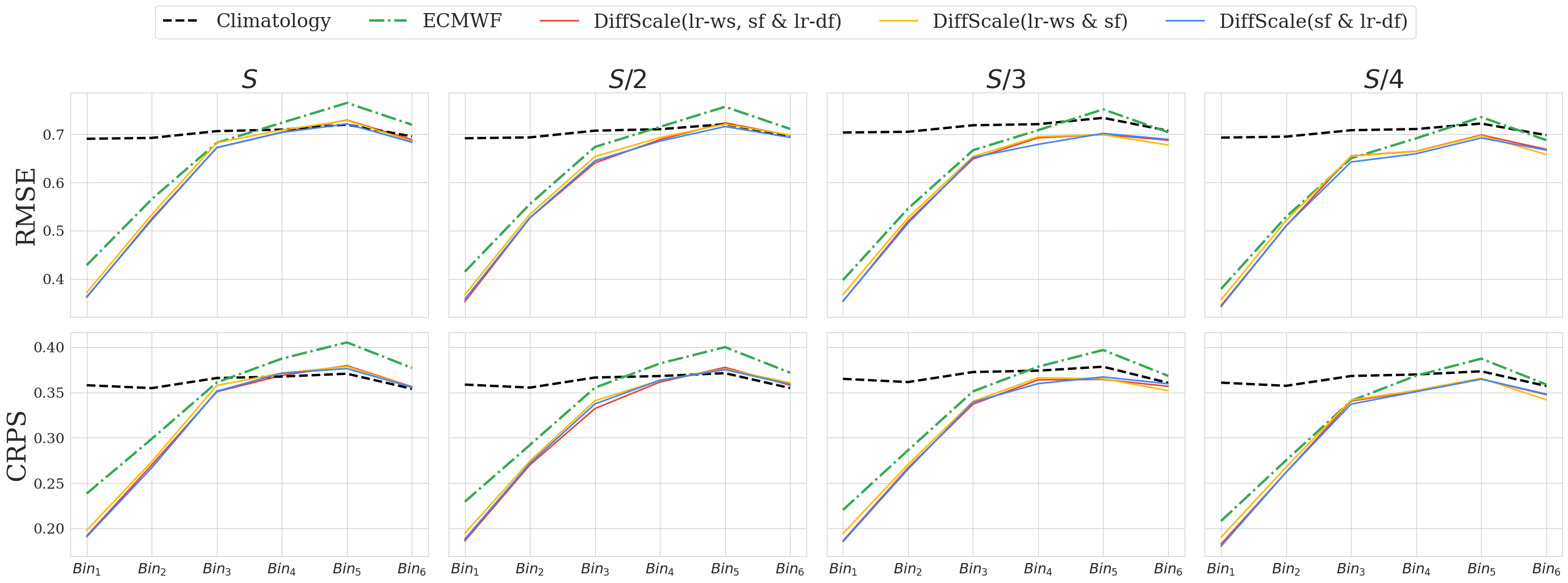}
    \caption{RMSE and CRPS metrics calculated across the different bins for each resolution. The different setups of \textit{DiffScale}'s experiments are shown as colored lines, while the ECMWF S2S model is represented as a blue line. The black line corresponds to climatology.}
    \label{fig:fig_eval_adx}
\end{figure*}

As stated in the main text, similar trends were observed in the root mean squared error (RMSE) and continuous ranked probability score (CRPS) metrics. \textit{DiffScale} outperforms the baseline ECMWF model across all lead times, suggesting its ability to capture the dynamics of the 10m wind speed (ws10), resulting in lower error rates. \Cref{fig:fig_eval_adx} depicts the performance of the models across all configuration setups, as assessed using the deterministic RMSE and probabilistic CRPS. \Cref{tab:RMSE,tab:CRPS} show the results we obtained for RMSE and CRPS, respectively.

\begin{table}[htbp]
    \centering
    \begin{scriptsize}
            \begin{tabular}{llcccccc}
                \toprule
                \multirow{2}{*}{} & \multirow{2}{*}{Method}      & \multicolumn{6}{c}{RMSE $\downarrow$}\\
                            \cmidrule{3-8}
                            &  & Bin$_1$ & Bin$_2$ & Bin$_3$ & Bin$_4$ & Bin$_5$ & Bin$_6$\\
                \midrule
                \multirow{5}{*}{$S$} 
                & Climatology                 & 0.691 & 0.693 & 0.707 & 0.710 & \bf{0.720} & 0.696  \\
                & ECMWF S2S                    &  0.429 & 0.566 & 0.682 & 0.724 & 0.765 & 0.720  \\
                \cdashlinelr{2-8}
                & Disffscale (sf \& lr-df)          & 0.421 & 0.557 & 0.690 & 0.712 & 0.737 &  0.690  \\
                & Disffscale (lr-ws, sf \& lr-df)   & 0.419 & 0.548 & 0.678 & \bf{0.696}  & 0.728 &  \bf{0.683}\\
                & Disffscale (lr-ws \& sf)          & \bf{0.363} & \bf{0.522} & \bf{0.673} & 0.704 & 0.722 & 0.684\\
                \midrule
                \multirow{5}{*}{$S/2$} 
                & Climatology                 & 0.692 &  0.694 & 0.708 & 0.711 & 0.721 & 0.697\\
                & ECMWF S2S                      & 0.415 & 0.555 & 0.674 & 0.716 & 0.757 & 0.711 \\
                \cdashlinelr{2-8}
                & Disffscale (sf \& lr-df)          & 0.420 & 0.566 & 0.664 & 0.701 & 0.727 & 0.694\\
                & Disffscale (lr-ws, sf \& lr-df)   & 0.412 & 0.550 & 0.659 & \bf{0.684} & 0.722 & \bf{0.689}  \\
                & Disffscale (lr-ws \& sf)          &  \bf{0.358} & \bf{0.527} & \bf{0.645} & 0.686 & \bf{0.716} & 0.690 \\
                \midrule
                \multirow{5}{*}{$S/3$} 
                & Climatology                 & 0.704 & 0.705 & 0.719 & 0.721 & 0.734 & 0.707 \\
                & ECMWF S2S                      & 0.398 & 0.546 & 0.667 & 0.709 & 0.752 & 0.704\\
                \cdashlinelr{2-8}
                & Disffscale (sf \& lr-df)          & 0.412 & 0.548 & 0.681 & 0.693 & 0.714 & \bf{0.678} \\
                & Disffscale (lr-ws, sf \& lr-df)   & 0.408 & 0.552 & 0.670 & 0.687 & \bf{0.702} & 0.680 \\
                & Disffscale (lr-ws \& sf)          & \bf{0.354} & \bf{0.516} & \bf{0.652} & \bf{0.679} & \bf{0.702} & 0.690\\
                \midrule
                \multirow{5}{*}{$S/4$} 
                & Climatology                 & 0.693 & 0.695 & 0.709 & 0.711 & 0.722 &  0.698  \\
                & ECMWF S2S                      &  0.379 & 0.529 & 0.650 & 0.692 & 0.736 & 0.688\\
                \cdashlinelr{2-8}
                & Disffscale (sf \& lr-df)          &  0.402 & 0.547 & 0.661 & 0.669 & 0.698 & \bf{0.658}\\
                & Disffscale (lr-ws, sf \& lr-df)   & 0.398 & 0.539 & 0.667 & 0.666 & 0.704 & \bf{0.658}\\
                & Disffscale (lr-ws \& sf)          & \bf{0.343} & \bf{0.510} & \bf{0.643} & \bf{0.660} & \bf{0.693} & 0.667\\
                \bottomrule
            \end{tabular}
    \end{scriptsize}
    \caption{RMSE scores. Reported are the best scores observed during training for respective bins.}
    \label{tab:RMSE}
\end{table}
\begin{table}[htbp]
    \centering
    \begin{scriptsize}
        \begin{tabular}{llcccccc}
            \toprule
            \multirow{2}{*}{} & \multirow{2}{*}{Method}      & \multicolumn{6}{c}{CRPS $\downarrow$}\\
                        \cmidrule{3-8}
                        &  & Bin$_1$ & Bin$_2$ & Bin$_3$ & Bin$_4$ & Bin$_5$ & Bin$_6$\\
            \midrule
            \multirow{5}{*}{$S$} 
            & Climatology                 & 0.358 & 0.355 & 0.366 & 0.367 & \bf{0.371} & \bf{0.354} \\
            & ECMWF S2S                  & 0.238 &  0.299 &  0.361 & 0.387 & 0.405 & 0.377  \\
                \cdashlinelr{2-8}
            & Disffscale (sf \& lr-df)          & 0.231 & 0.295 & 0.363 & 0.374 & 0.384 & 0.358 \\
            & Disffscale (lr-ws, sf \& lr-df)      & 0.229 & 0.288 & 0.356 & \bf{0.364} &  0.379 & \bf{0.354} \\
            & Disffscale (lr-ws \& sf)           & \bf{0.191} & \bf{0.267} & \bf{0.351} & 0.371 & 0.376 &  0.356 \\
            \midrule
            \multirow{5}{*}{$S/2$} 
            & Climatology                 & 0.359 & 0.355 & 0.366 & 0.368 & \bf{0.371} & 0.354 \\
            & ECMWF S2S                     & 0.229 & 0.292 & 0.355 & 0.382 & 0.371 &  0.355 \\
                \cdashlinelr{2-8}
            & Disffscale (sf \& lr-df)          & 0.230 &  0.299 & 0.345 &  0.368 &  0.380 & 0.359 \\
            & Disffscale (lr-ws, sf \& lr-df)      &  0.227 & 0.290 & 0.344 & \bf{0.360} & 0.377 &  \bf{0.353} \\
            & Disffscale (lr-ws \& sf)           & \bf{0.188} & \bf{0.272} & \bf{0.337} & 0.363 & 0.375 &  0.359 \\
            \midrule
            \multirow{5}{*}{$S/3$} 
            & Climatology                 & 0.358 & 0.355 & 0.366 & 0.368 & 0.371 & 0.354 \\
            & ECMWF S2S                      & 0.220 & 0.286 &  0.351 & 0.378 & 0.397 & 0.368 \\
                \cdashlinelr{2-8}
            & Disffscale (sf \& lr-df)          & 0.224 & 0.288 & 0.355 & 0.364 &  0.374 & \bf{0.352}  \\
            & Disffscale (lr-ws, sf \& lr-df)      & 0.224 & 0.288 & 0.350 & 0.363 & \bf{0.367} &  0.353 \\
            & Disffscale (lr-ws \& sf)           & \bf{0.185} & \bf{0.265} & \bf{0.339} & \bf{0.360} &  \bf{0.367} & 0.359 \\
            \midrule
            \multirow{5}{*}{$S/4$}
            & Climatology                 & 0.361 & 0.357 & 0.368 & 0.370 & 0.373 & 0.357 \\
            & ECMWF S2S                      & 0.208 & 0.275 &  \bf{0.341} &  0.368 & 0.387 & 0.359  \\
                \cdashlinelr{2-8}
            & Disffscale (sf \& lr-df)          &  0.220 & 0.289 & 0.347 & \bf{0.354} &  \bf{0.366} & \bf{0.342}   \\
            & Disffscale (lr-ws, sf \& lr-df)   & 0.218 & 0.283 &  0.348 &  \bf{0.354} & 0.367 & \bf{0.342}   \\
            & Disffscale (lr-ws \& sf)          & \bf{0.191} & \bf{0.267} & 0.352 & 0.371 & 0.376 &  0.356 \\
            \bottomrule
        \end{tabular}
    \end{scriptsize}
    \caption{CRPS scores. Reported are the best scores observed during training for respective bins.}
     \label{tab:CRPS}
\end{table}

\section{Qualitative assessment}  \label{sup:qualitative}
In the following, we will depict the spatial evaluation of the two model configurations \textit{lr-ws}, \textit{sf} \& \textit{lr-df}(low-resolution ECMWF wind-speed forecast, static fields and low-resolution dynamic fields), and \textit{sf} \& \textit{lr-df} (static fields and low-resolution dynamic fields). \Cref{fig:figbias-ids,fig:figacc-ids,fig:figmae-ids,fig:figcrps-ids} depict the bias, anomaly correlation coefficient (ACC), mean absolute error (MAE), and CRPS for the \textit{lr-ws}, \textit{sf} \& \textit{lr-df} configuration and \Cref{fig:figbias-ds,fig:figacc-ds,fig:figmae-ds,fig:figcrps-ds} depict the bias, anomaly correlation coefficient, MAE and CRPS score for the \textit{sf} \& \textit{lr-df} configuration. 

All figures depict the performance of \textit{DiffScale} w.r.t. to interpolated ECMWF ws10 forecasts. For all configurations and scores, \textit{DiffScale} shows improvements, particularly in coastal regions and areas with high elevation. Among all \textit{DiffScale} configurations, the best-performing setup is \textit{lr-ws \& sf}, as reported in the main text. Performance differences arising from dynamic fields as inputs are marginal. Even without the low resolution of ws10 forecasts, the model outperforms the interpolated forecasts, highlighting the advantages of using a generative model based on probabilistic principles for climate variable predictions.

\input{supplementary_spatial_input_dyn_static}
\input{supplementary_spatial_dyn_static}

%% file: supplementary_spatial_input_dyn_static.tex

\begin{figure*}[p]
    \centering
    \includegraphics[width=0.925\textwidth]{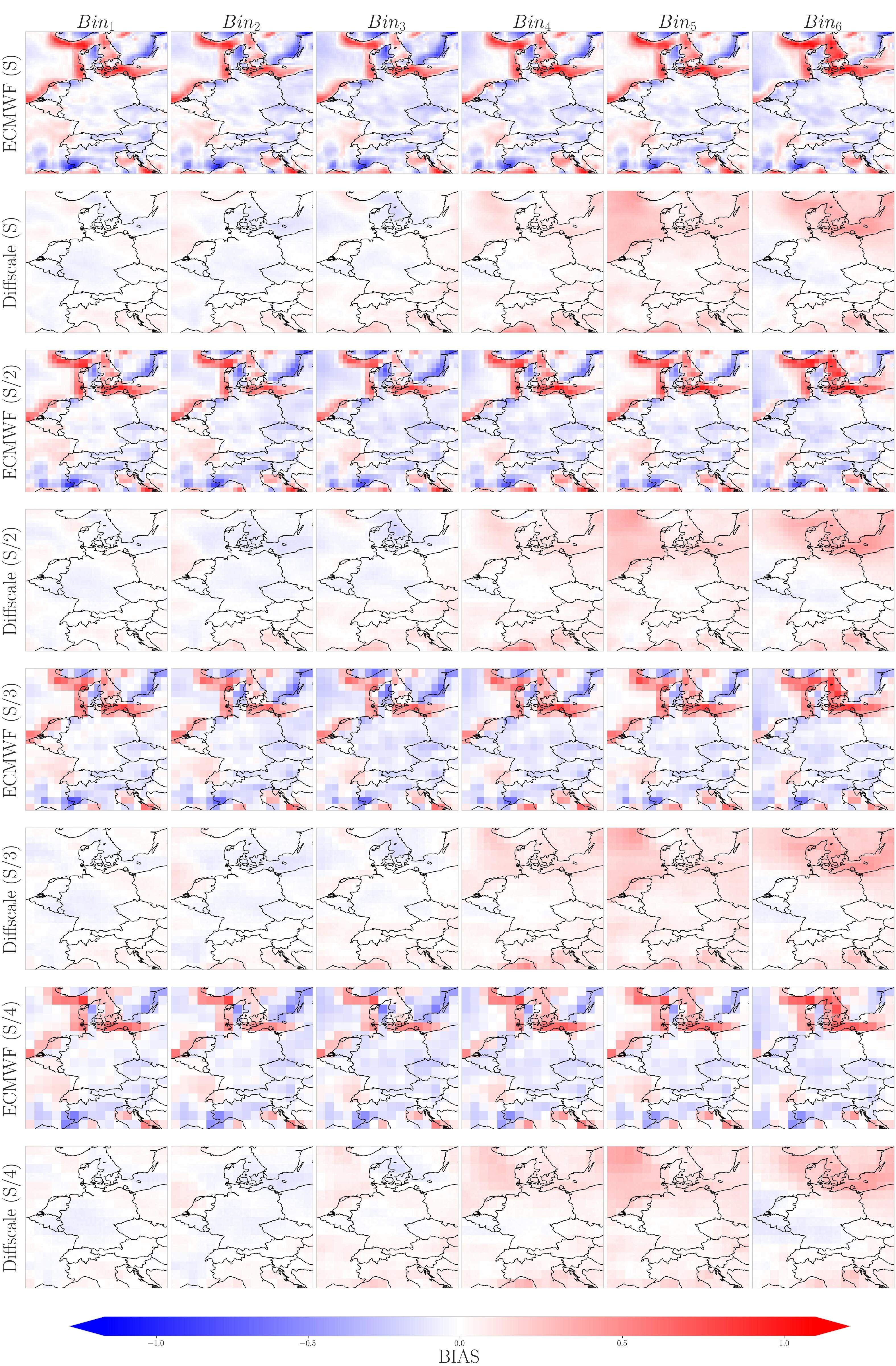}
    \caption{Bias of ws10m (m/s) calculated at each spatial resolution and lead time bin for the ECMWF and \textit{DiffScale} (lr-ws, sf \& lr-df). }
    \label{fig:figbias-ids}
\end{figure*}

\begin{figure*}[p]
    \centering
    \includegraphics[width=0.925\textwidth]{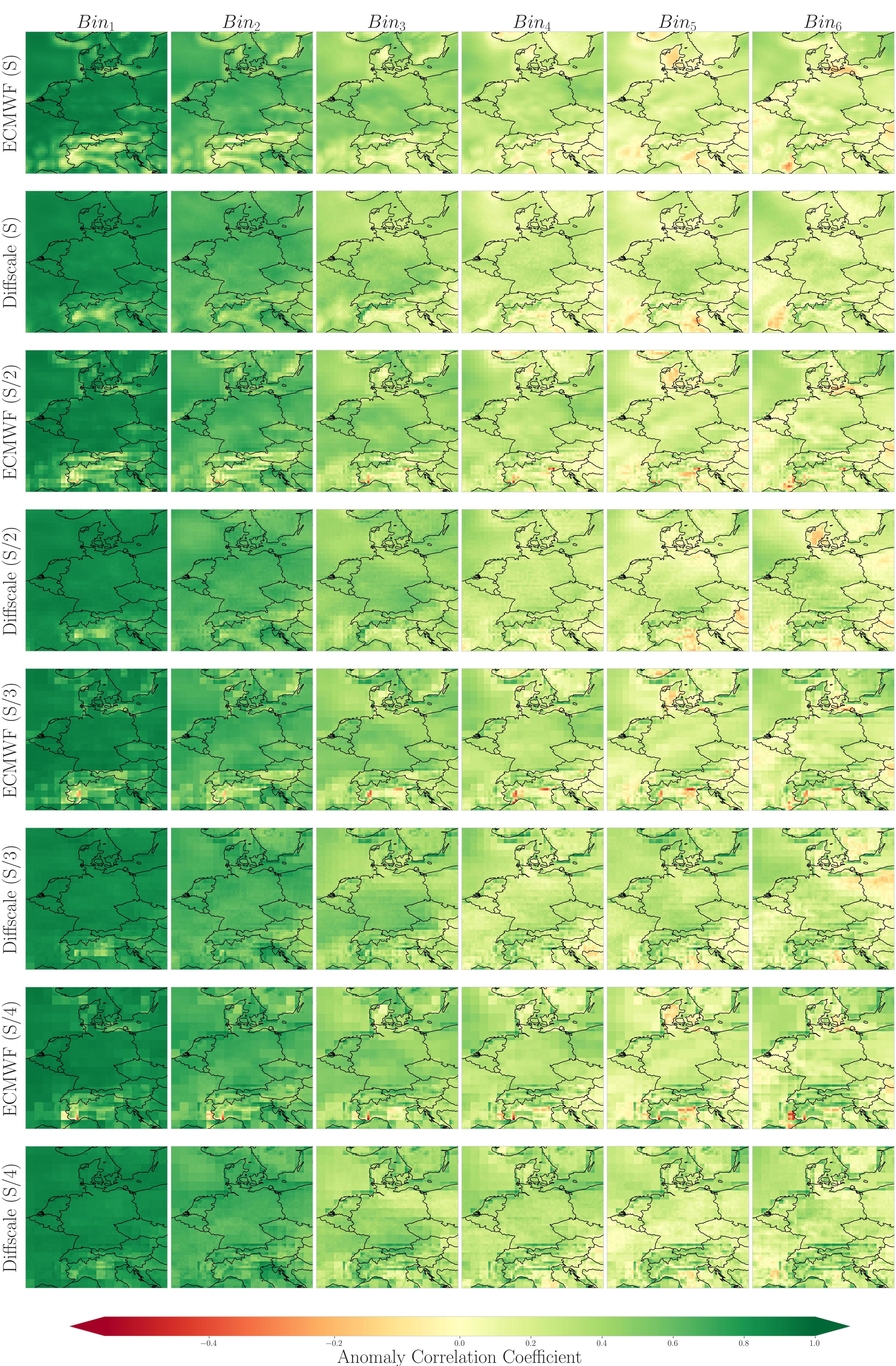}
    \caption{Spatial distribution of the anomaly correlation coefficient (ACC) for the ECMWF and the \textit{DiffScale} (lr-ws, sf  \& lr-df)  model for each resolution and lead time bins. Note that the ACC are calculated considering all forecast times. }
    \label{fig:figacc-ids}
\end{figure*}

\begin{figure*}[p]
    \centering
    \includegraphics[width=0.925\textwidth]{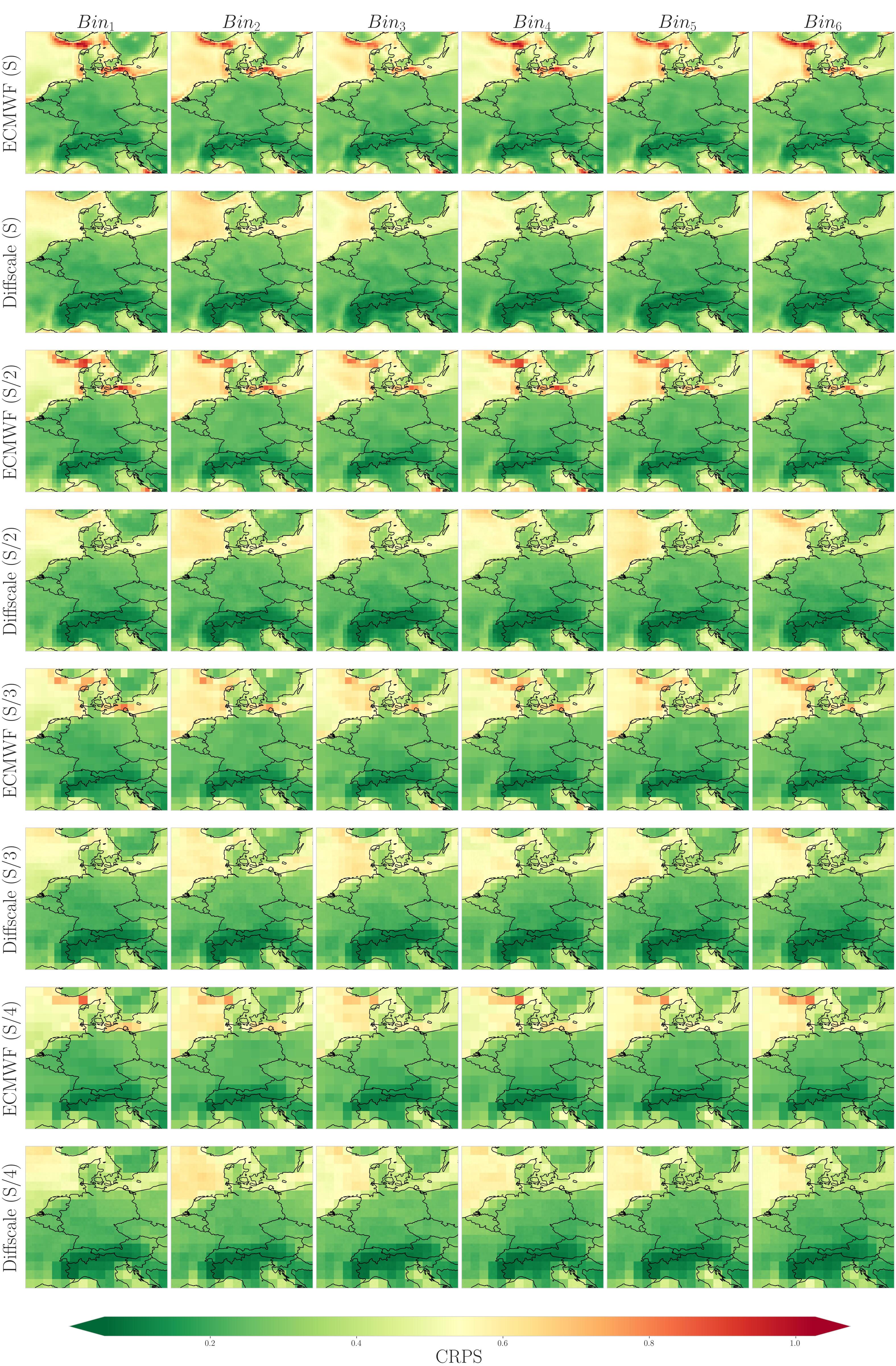}
    \caption{Spatial distribution of the CRPS for the ECMWF and the \textit{DiffScale} (lr-ws, sf \& lr-df)  model for each resolution and lead time bins.}
    \label{fig:figcrps-ids}
\end{figure*}

\begin{figure*}[p]
    \centering
    \includegraphics[width=0.925\textwidth]{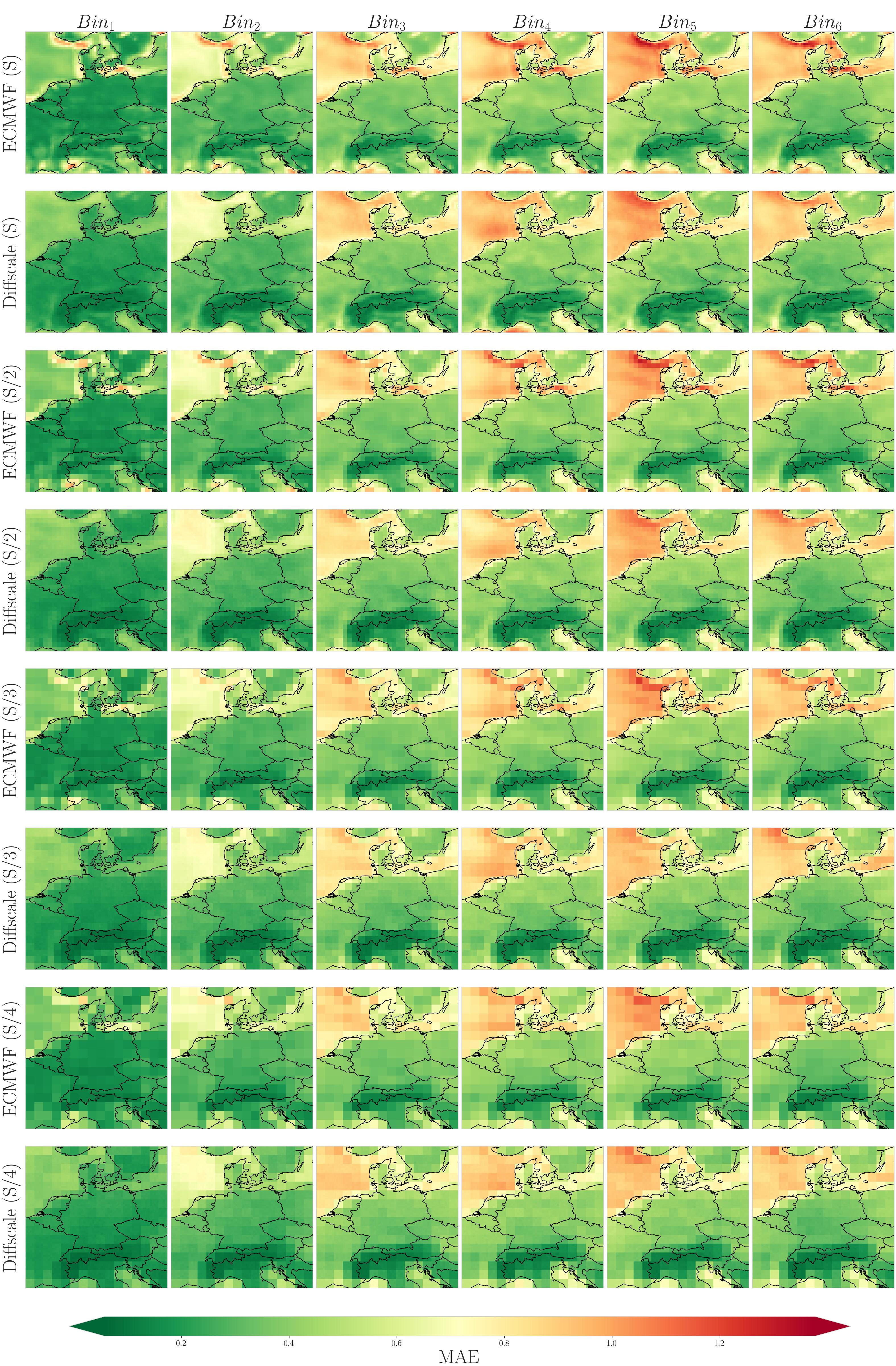}
    \caption{Spatial distribution of the MAE for the ECMWF and the \textit{DiffScale} (lr-ws, sf  \& lr-df)  model for each resolution and lead time bins.}
    \label{fig:figmae-ids}
\end{figure*}

%% file: supplementary_spatial_dyn_static.tex

\begin{figure*}[p]
    \centering
    \includegraphics[width=0.925\textwidth]{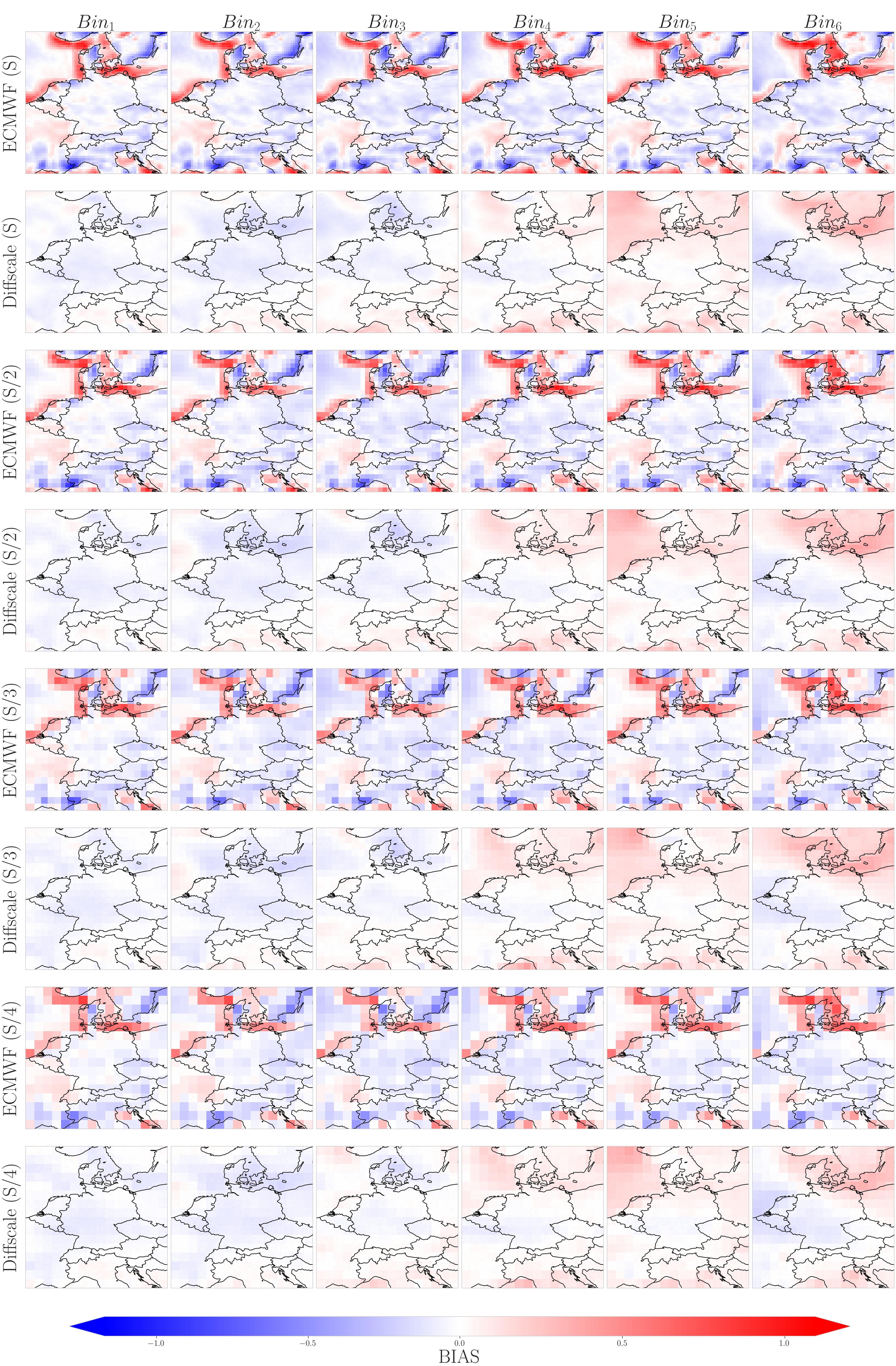}
    \caption{Bias of ws10m (m/s) calculated at each spatial resolution and lead time bin for the ECMWF and \textit{DiffScale} (sf \& lr-df). }
    \label{fig:figbias-ds}
\end{figure*}

\begin{figure*}[p]
    \centering
    \includegraphics[width=0.925\textwidth]{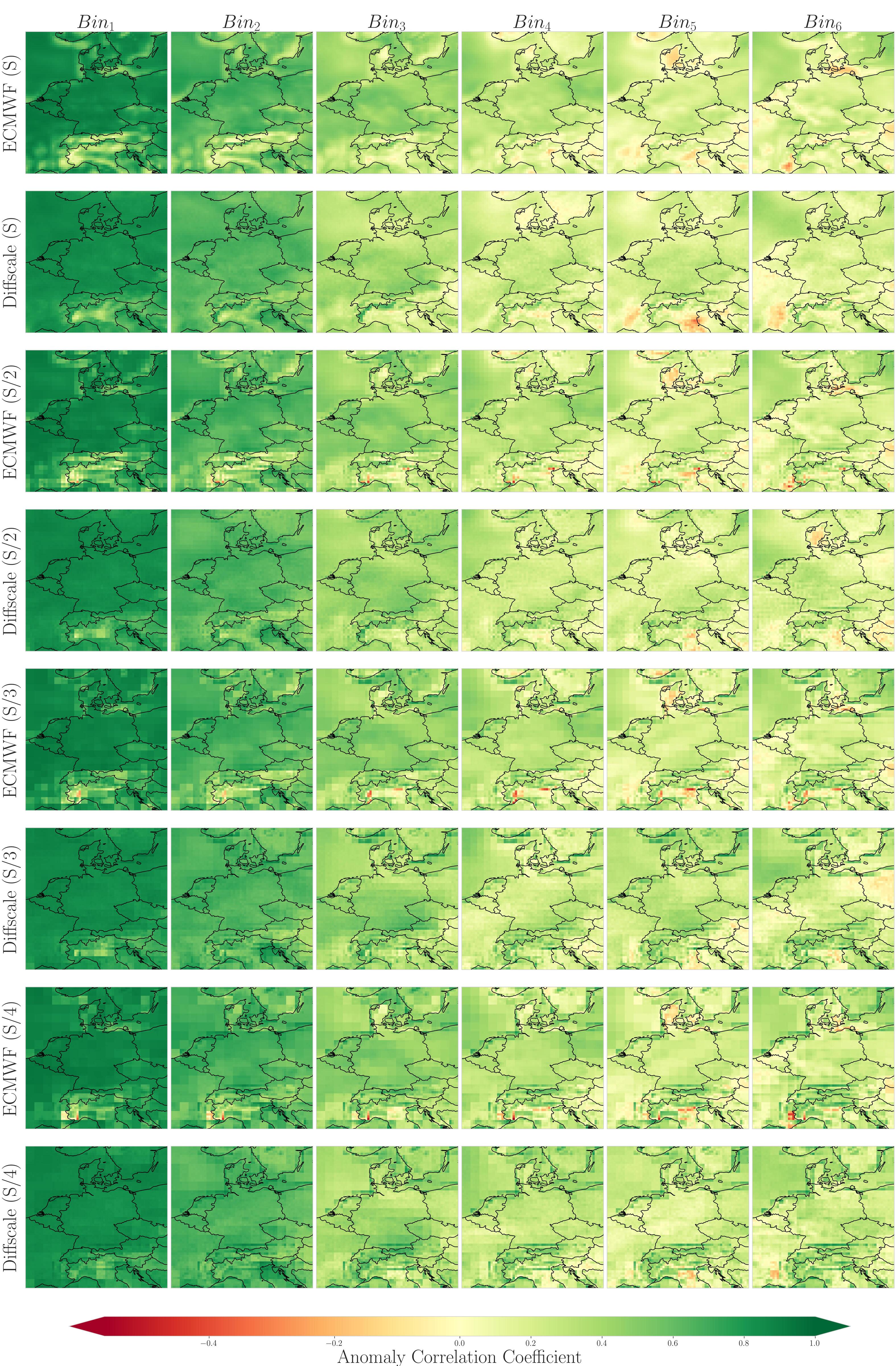}
    \caption{Spatial distribution of the anomaly correlation coefficient (ACC) for the ECMWF and the \textit{DiffScale} (sf \& lr-df)  model for each resolution and lead time bins. Note that the ACC are calculated considering all forecast times. }
    \label{fig:figacc-ds}
\end{figure*}

\begin{figure*}[p]
    \centering
    \includegraphics[width=0.925\textwidth]{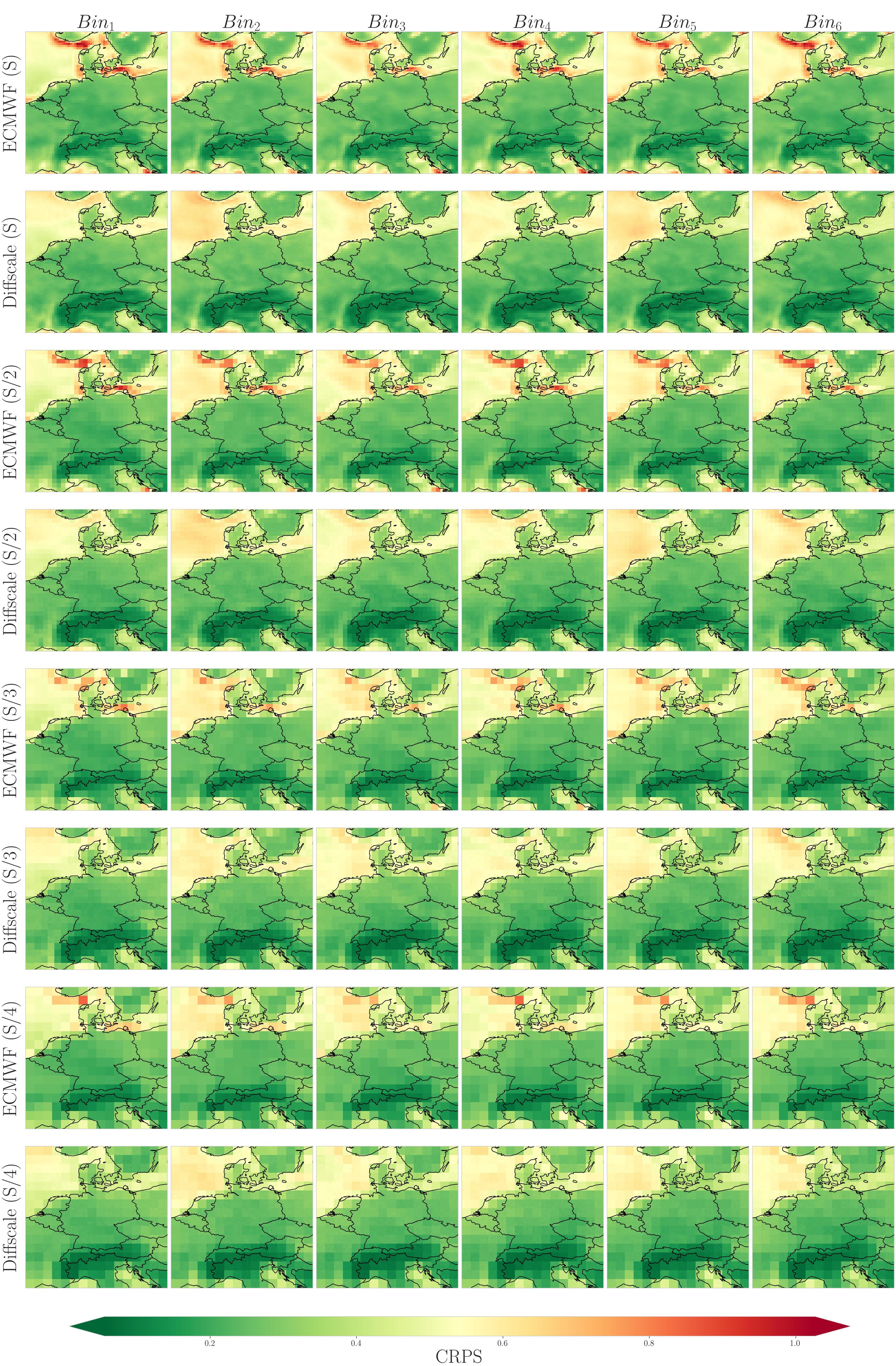}
    \caption{Spatial distribution of the CRPS for the ECMWF and the \textit{DiffScale} (sf \& lr-df)  model for each resolution and lead time bins.}
    \label{fig:figcrps-ds}
\end{figure*}

\begin{figure*}[p]
    \centering
    \includegraphics[width=0.925\textwidth]{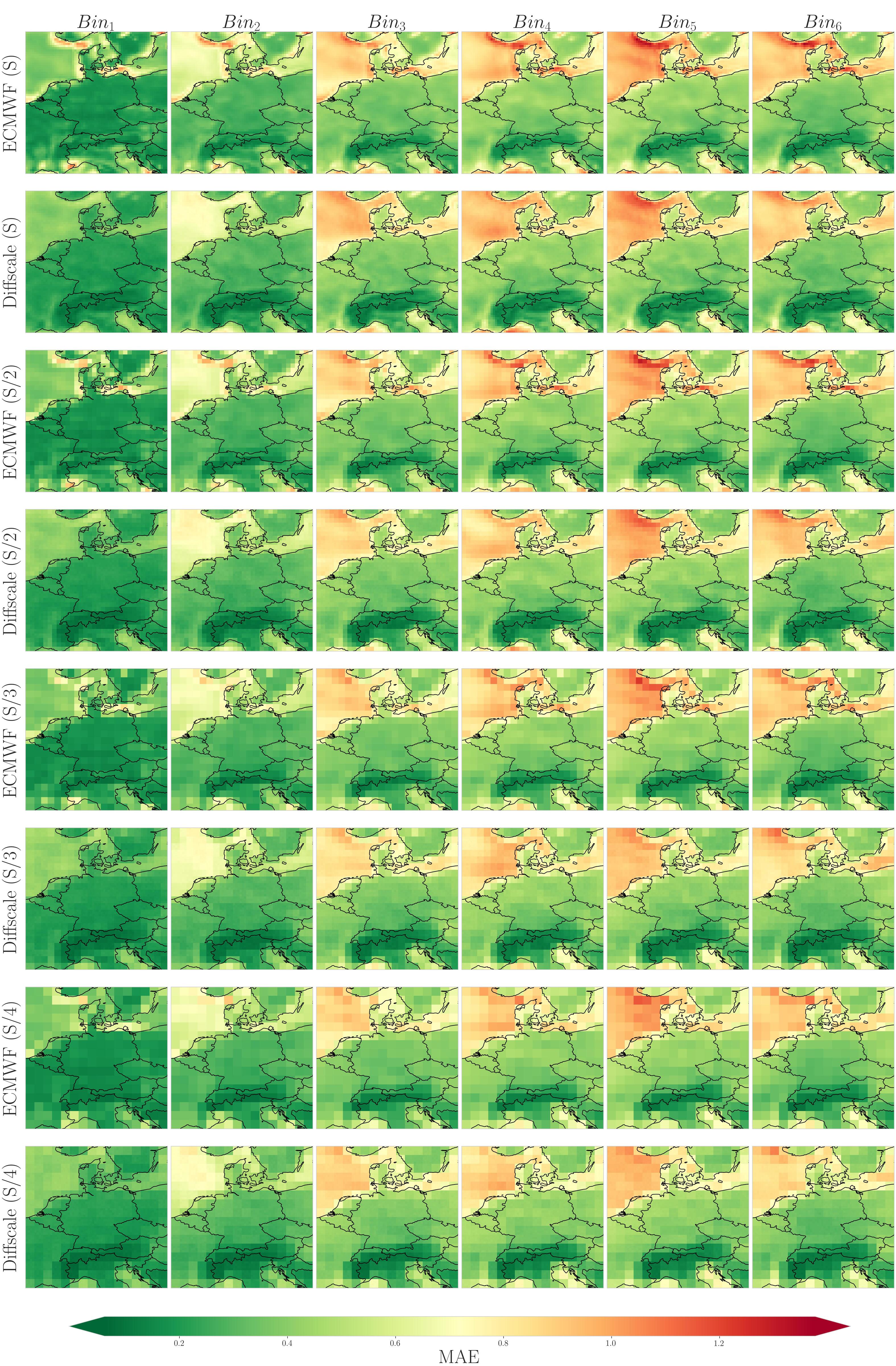}
    \caption{Spatial distribution of the MAE for the ECMWF and the \textit{DiffScale} (sf \& lr-df) model for each resolution and lead time bins.}
    \label{fig:figmae-ds}
\end{figure*}